%% file: iclr2026_conference.tex
\documentclass{article} 
\usepackage{iclr2026_conference,times}

\input{math_commands.tex}

\usepackage{hyperref}
\usepackage{url}
\usepackage{booktabs}
\usepackage{xcolor}
\usepackage{colortbl}
\usepackage{graphicx}
\usepackage{tcolorbox}
\usepackage{algorithm}
\usepackage{algorithmic}

\newtheorem{assumption}{Assumption}

\title{Unifying Dynamic Tool Creation and Cross-Task Experience Sharing through Cognitive Memory Architecture}

\author{Jiarun Liu, Shiyue Xu, Yang Li, Shangkun Liu, Yongli Yu, Caopeng\\
\texttt{\{liujiarun.1, xushiyue.6, liyang1236\}@jd.com}\\
}

\iclrfinalcopy 
\begin{document}

\maketitle

\begin{abstract}
Large Language Model agents face fundamental challenges in adapting to novel tasks due to limitations in tool availability and experience reuse. Existing approaches either rely on predefined tools with limited coverage or build tools from scratch without leveraging past experiences, leading to inefficient exploration and suboptimal performance. We introduce SMITH (Shared Memory Integrated Tool Hub), a unified cognitive architecture that seamlessly integrates dynamic tool creation with cross-task experience sharing through hierarchical memory organization. SMITH organizes agent memory into procedural, semantic, and episodic components, enabling systematic capability expansion while preserving successful execution patterns. Our approach formalizes tool creation as iterative code generation within controlled sandbox environments and experience sharing through episodic memory retrieval with semantic similarity matching. We further propose a curriculum learning strategy based on agent-ensemble difficulty re-estimation. Extensive experiments on the GAIA benchmark demonstrate SMITH's effectiveness, achieving 81.8\% Pass@1 accuracy and outperforming state-of-the-art baselines including Alita (75.2\%) and Memento (70.9\%). Our work establishes a foundation for building truly adaptive agents that continuously evolve their capabilities through principled integration of tool creation and experience accumulation.
\end{abstract}

\begin{figure}[htbp]
    \centering
    \includegraphics[width=0.87\textwidth]{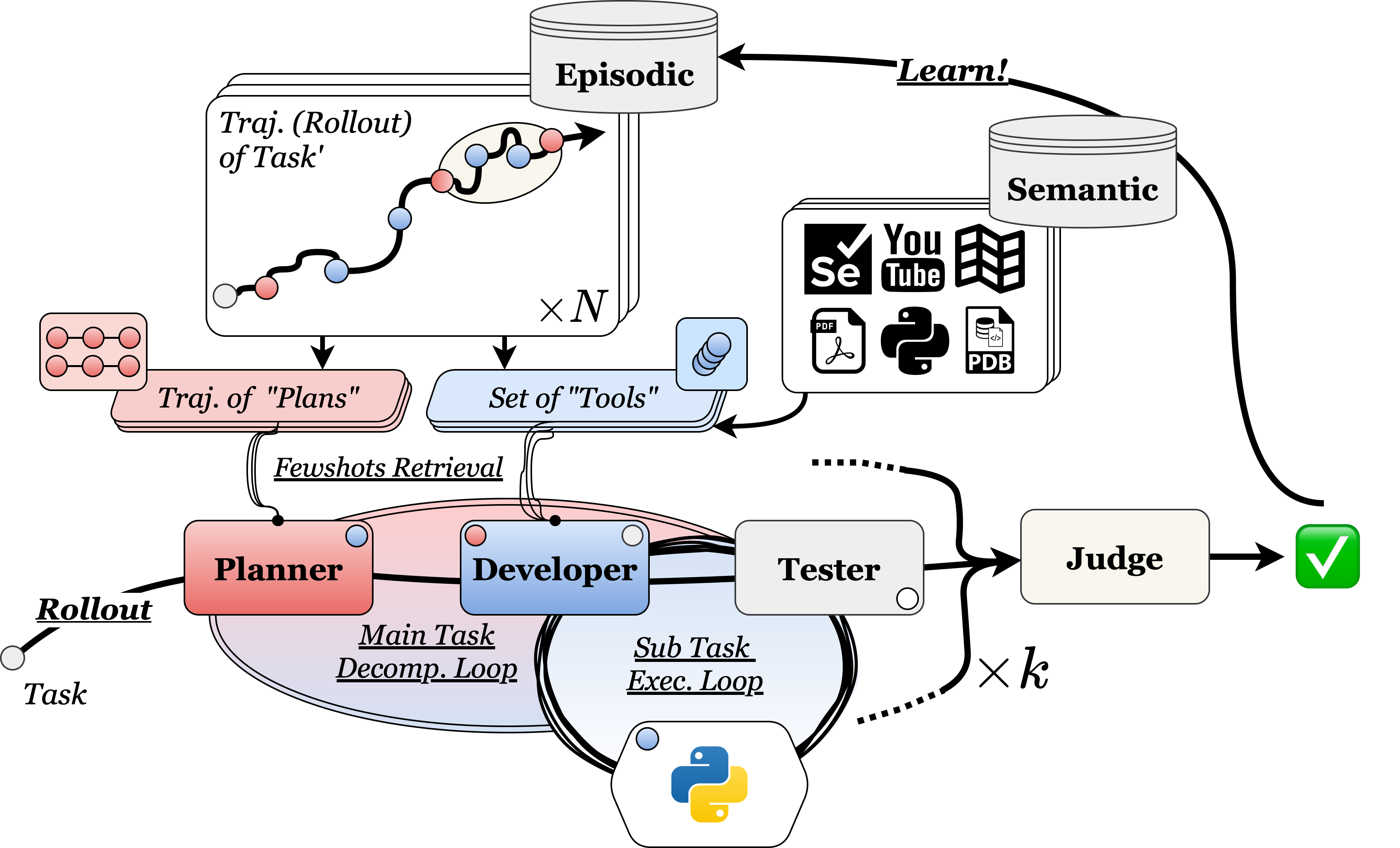}
    \caption{SMITH architecture overview. Each agent rollout involves two nested loops: an inner developer-tester loop for iterative code generation and debugging, and an outer planner loop for sub-plan execution. $\times k = 3$ represents 3-path sampling with LLM-as-a-judge consensus voting. Upon successful task completion, corresponding experiences are processed, embedded, and stored for future learning and reuse.}
    \label{fig:arch}
\end{figure}

\section{Introduction}
\label{intro}

The development of general AI assistants capable of tackling diverse, real-world tasks remains a fundamental challenge in artificial intelligence. While Large Language Models (LLM) have demonstrated remarkable reasoning capabilities, their application to complex problem-solving scenarios is often limited by two critical bottlenecks: the availability of appropriate tools for task execution and the ability to leverage past experiences for novel situations. Current approaches address these challenges in isolation—tool learning frameworks like Toolformer \citep{schick2023toolformer} rely on predefined tool collections with limited coverage, while recent tool creation methods such as Alita \citep{qiu2025alita} generate tools from scratch without systematic reuse. Similarly, experience sharing approaches like Memento \citep{zhou2025memento} focus on cross-task memory transfer but lack integrated tool creation capabilities. This fragmentation prevents agents from achieving the adaptive, cumulative learning characteristic of human problem-solving, where tools are created, refined, and reused across related tasks while successful strategies are systematically transferred to new domains.

We propose SMITH (Shared Memory Integrated Tool Hub), a unified cognitive architecture that bridges this gap by seamlessly integrating dynamic tool creation with cross-task experience sharing through a hierarchical memory framework. Drawing inspiration from cognitive architectures for language agents \citep{sumers2023cognitive}, SMITH organizes agent memory into procedural, semantic, and episodic components, enabling systematic capability expansion while preserving successful execution patterns across tasks. Our approach formalizes tool creation as an iterative code generation process within controlled sandbox environments, and experience sharing through episodic memory retrieval with semantic similarity matching. To optimize learning efficiency, we introduce a novel curriculum learning strategy based on agent-ensemble difficulty re-estimation that reranks tasks according to agent-specific capability assessments rather than human annotations. This unified framework enables agents to continuously evolve their problem-solving capabilities through principled integration of tool creation and experience accumulation, establishing a foundation for truly adaptive AI systems that can tackle the complexity and diversity of real-world challenges.

\section{Related Work}
\label{related_work}

\textbf{Multi-Agent Systems and General AI Assistants.} Benchmarks like GAIA \citep{mialon2023gaia} evaluate general AI assistants through real-world questions requiring reasoning, multi-modality handling, and tool-use proficiency. AutoAgent \citep{tang2025autoagent} democratizes development through zero-code interfaces, OWL \citep{hu2025owl} enables cross-domain adaptation via hierarchical architectures, and AWorld \citep{yu2025aworld} accelerates experience collection by 14.6× through distributed infrastructure. These approaches establish foundations for scalable, general-purpose AI assistants.

\textbf{Memory Architectures for Language Agents.} Context window limitations have driven extensive research into memory architectures for language agents. Building on memory networks \citep{weston2014memory} and retrieval-augmented generation \citep{lewis2020retrieval}, \citet{sumers2023cognitive} established theoretical foundations through Cognitive Architectures for LLM Agents, organizing agent memory into working, episodic, semantic, and procedural memory hierarchies. Practical implementations include MemGPT \citep{packer2023memgpt} with OS-inspired virtual context management, Mem0 \citep{chhikara2025mem0} with graph-based representations, and self-controlled frameworks \citep{wang2023enhancing} achieving 77.1\% accuracy with 91\% lower latency. These advances enable persistent, context-aware systems like Generative Agents \citep{park2023generative} and ReAct \citep{yao2022react}.

\textbf{Tool Learning and Tool Creation.} While tool learning utilizes pre-existing tools \citep{schick2023toolformer}, it requires human developers to design tools beforehand. Recent work shifted toward tool creation, enabling autonomous tool generation at runtime. Early methods like CRAFT \citep{yuan2024craft}, CREATOR \citep{qian2023creator}, and LATM \citep{cai2024toolmakers} generated simple Python functions but lacked system interaction capabilities. Advanced frameworks expanded these capabilities: \citet{wolflein2025llm} introduced ToolMaker for transforming scientific repositories into LLM-compatible tools, while \citet{qiu2025alita} proposed Alita achieving 75.15\% on GAIA through ``minimal predefinition and maximal self-evolution'' using Model Context Protocols. The key innovation, emerging from frameworks like SmolAgent \citep{smolagents}, is the ability to save and cache generated tools, forming closed-loop systems where successful tools become reusable assets.

\textbf{Experience Sharing.} Parameter-based methods like AWorld \citep{yu2025aworld} and WebShaper \citep{tao2025webshaper} employ supervised fine-tuning followed by reinforcement learning but suffer from unclear memory hierarchies and inability to perform continual learning during inference. Memory-based approaches address these limitations by storing task execution memories as episodic traces without parameter modifications. \citet{zhou2025memento} introduced Memento with Memory-augmented Markov Decision Process (M-MDP) achieving 87.88\% Pass@3 on GAIA, \citet{li2025cross} proposed MAEL for multi-agent cross-task experiential learning, and \citet{yang2024cops} developed CoPS with pessimism-based experience selection. As noted in cognitive architectures \citep{sumers2023cognitive}, experience sharing manages episodic memory through embedding-based retrieval systems with structural commonalities to memory management frameworks.

\section{Method}
\label{method}

\subsection{Formalization of Dynamic Tool Creation}
\label{tool_creaction}

We formalize the dynamic tool creation process as an interactive code generation and refinement procedure within a controlled execution environment. This formalization captures the iterative nature of tool development, where agents continuously write, test, debug, and refine code until successful tool implementation is achieved.

\textbf{Sandbox Environment and Agent Interaction.} We define a python sandbox execution environment $ \langle \mathcal{E}, \texttt{exec}, \texttt{feedback} \rangle$ where $\mathcal{E}$ represents the current environment state, $\texttt{exec}: \mathcal{E} \times C \rightarrow \mathcal{E} \times O$ executes code $C$ and returns updated state and output $O$, and $\texttt{feedback}: O \rightarrow F$ provides structured error or success feedback $F$. Let agent $a$ represent the code-writing entity that interacts with environment through an iterative debugging loop. At each iteration $t$, the agent maintains code $c_t$ and receives feedback $f_t$ from the sandbox environment.

\textbf{Interactive Tool Creation Process.} Given a task specification $\tau$, the tool creation process unfolds as an iterative refinement sequence
\begin{equation}
c_{t+1} = \texttt{agent}(c_t, f_t, \tau, \mathcal{C}_{\text{code}})
\end{equation}
where $\mathcal{C}_{\text{code}} = \{(c_{t-1}, f_{t-1}), (c_{t-2}, f_{t-2}), \ldots\}$ represents contextual memory containing historical code-feedback pairs, debugging patterns, and successful implementation trajectories from previous iterations. The process continues until the sandbox environment returns successful feedback
\begin{equation}
f_t = \texttt{feedback}(\texttt{exec}(\mathcal{E}_t, c_t)) \in \{\checkmark, e_t\}
\end{equation}
When $f_t = e_t$, the agent analyzes the error $e_t$ and generates refined code $c_{t+1}$. This debug-and-refine cycle continues until $f_t = \checkmark$. Upon successful execution, typically the code $c_\text{done}$ undergoes encapsulation to form a \textit{tool}, or from a more comprehensive perspective, it can be formalized as a tool creation memory episode where the concept of \textit{tool} is dissolved into past action execution
\begin{equation}
\label{eq3}
    T = \{ c_{done}, (c_{0}, \mathcal{E}_0, f_0) ... (c_\text{done}, \mathcal{E}_\text{done}, \checkmark) \}
\end{equation}
where $(c_{0}, \mathcal{E}_0, f_0) \rightarrow (c_\text{done}, \mathcal{E}_\text{done}, \checkmark)$ captures the complete debugging trajectory. The tool repository $\mathbb{T}$ evolves dynamically as $ \mathbb{T} \leftarrow \mathbb{T} \cup \{T\}$, enabling future tool reuse and composition.

\subsection{Formalization of Cross-Task Experience Sharing}
\label{cross-task}
We formalize cross-task experience sharing through an episodic memory framework that enables agents to leverage previous successful execution patterns with semantically similarity. We establish the following assumption with $\mathcal{T} = \{\tau_1, \tau_2, \ldots, \tau_n\}$ denoting the task universe.
\begin{assumption}[Semantic Task Similarity]
\label{assumption1}
Two tasks $\tau_i, \tau_j \in \mathcal{T}$ are considered semantically similar if their problem structures and solution requirements exhibit similar patterns in semantic space, as measured by the similarity of their embedding representations $\Phi(\tau_i)$ and $\Phi(\tau_j)$. Formally, we define semantic similarity as $\text{sim}(\Phi(\tau_i), \Phi(\tau_j)) > \theta$ for some threshold $\theta$. Tasks satisfying this similarity criterion enable transferability of execution experiences across these tasks.
\end{assumption}
Each agent $j$ maintains its own episodic memory $\mathcal{M}_{\text{ep}}^{(j)} = \{e_1^{(j)}, e_2^{(j)}, \ldots, e_k^{(j)}\}$, where each experience encapsulates a complete trajectory
\begin{equation}
\label{eq4}
e_i^{(j)} = \{ \tau_l, (s_0, a_0) ... (s_\text{done}, a_\text{done}) \}
\end{equation}
where $s_t$ represents the agent's observation state at step $t$ (including task context, current progress, and environmental feedback), and $a_t$ denotes the action taken (e.g., code generation, tool invocation, or sub-plan decomposition). 

We then define abstraction function $\Phi:e_i^{(j)} \rightarrow \mathbf{m}_i$ that varies by action space, where code writing actions require summarization before embedding, while planning agents perform intention decomposition and augmentation on proposed plans (details in Sec. \ref{implementation}).

\textbf{Experience Retrieval and Policy Enhancement.} Given current task $\tau$ and state $s_t$, agent node $j$ retrieves top-$k$ experiences via similarity scoring 
\begin{equation}
r(e_i^{(j)}, \tau, s_t) = \langle \Phi(\{ \tau, (s_t, \cdot) \}), \mathbf{m}_i \rangle
\end{equation}
The top-$k$ experiences are retrieved as
\begin{equation}
m_t = \texttt{TopK}_{e_i^{(j)} \in \mathcal{M}_{\text{ep}}^{(j)}} r(e_i^{(j)}, \tau, s_t)
\end{equation}
and actions are sampled as $a_t \sim \pi(\tau \oplus s_t \oplus m_t)$. 

\textbf{Memory Update and Experience Accumulation.} Upon successful task completion, the complete execution trajectory is added to the agent's episodic memory repository $\mathcal{M}_{\text{ep}}$, with the corresponding semantic representation computed via the abstraction function $\Phi$ for efficient future retrieval.

The formulation in Eq. \ref{eq4} exhibits \textit{structural duality} with tool creation framework from Sec. \ref{tool_creaction}. When action $a_t$ corresponds to code segment $c_t$, Eq. \ref{eq4} and Eq. \ref{eq3} demonstrate fundamental equivalence, which motivates us to construct a unified framework from a holistic perspective.

\subsection{Unified Cognitive Memory Architecture}
\label{framework}

Existing agent development approaches fail to integrate tool creation and experience sharing due to inadequate memory management frameworks. Current methods either rely on predefined tool collections with limited coverage or build tools from scratch, which is computationally expensive and restricts exploration \citep{qiu2025alita}. We propose a unified cognitive architecture, namely SMITH (Shared Memory Integrated Tool Hub), that seamlessly integrates dynamic tool creation with cross-task episodic learning.

\textbf{Hierarchical Memory Organization.} Drawing inspiration from cognitive architectures for language agents \citep{sumers2023cognitive}, SMITH organizes agent memory into a structured hierarchy that enables modular agent design and sophisticated decision-making procedures
\begin{equation}
\mathcal{M} = \{ \mathcal{M}_{\text{proc}}, \{\mathcal{M}_{\text{sem}}, \mathcal{M}_{\text{ep}}\} \}
\end{equation}
where each memory component serves distinct but complementary functions in the agent's reasoning process. \textbf{Procedural Memory ($\mathcal{M}_{\text{proc}}$)} encapsulates the agent's fundamental operational knowledge, including system prompts, behavioral guidelines, and the implicit knowledge encoded in LLM parameters $\Theta$. This memory component remains relatively static and provides the foundational reasoning capabilities that guide agent behavior across all tasks. \textbf{Semantic Memory ($\mathcal{M}_{\text{sem}}$)} contains externally provided knowledge and demonstrations, including human-crafted tool examples, transfer learning experiences from related task domains, and initial few-shot demonstrations. This memory serves as the bridge between human expertise and agent capabilities, providing high-quality starting points for tool creation and task execution. \textbf{Episodic Memory ($\mathcal{M}_{\text{ep}}$)} stores online task execution experiences as formalized in Sec. \ref{cross-task}, enabling continuous learning and adaptation through accumulated problem-solving patterns.

The overall memory-augmented decision process integrates all memory components through a unified retrieval and application mechanism
\begin{equation}
a_t \sim \pi(\tau \oplus s_t \oplus \texttt{Retrieve}(\mathcal{M}_{\text{sem}} \cup \mathcal{M}_{\text{ep}}, \tau, s_t) \mid \mathcal{M}_{\text{proc}})
\end{equation}
where $\texttt{Retrieve}$ accesses both $\mathcal{M}_{\text{ep}}$ and $\mathcal{M}_{\text{sem}}$ repositories using consistent similarity-based scoring, and $\mathcal{M}_{\text{proc}}$ provides the foundational reasoning context. Note that SMITH applies not only to coding agents that create executable tools, but also to higher-level entities such as planning agents whose actions consist of sub-intentions and strategic decompositions.

\textbf{Unified Memory Integration.} Both semantic and episodic memories maintain equivalent granularity with dense embedding representations $\mathbf{m}$, enabling seamless integration within a unified retrieval framework that supports elegant scalability and modular agent development.

\subsection{Model-based Difficulty Re-estimation for Curriculum Learning}
\label{curriculum}

The unified memory architecture in SMITH naturally motivates a \textit{curriculum learning} approach. Since agents can retrieve experiences from semantically similar prior tasks, we hypothesize that strategic task ordering can maximize the effectiveness of cross-task experience transfer.

\begin{assumption}[Task Dependency for Curriculum Learning]
\label{assumption2}
For any task $\tau_i \in \mathcal{T}$, there exists a finite set of prerequisite tasks $\mathcal{P}(\tau_i) = \{\tau_{j_1}, \tau_{j_2}, \ldots, \tau_{j_k}\} \subseteq \mathcal{T}$ such that successful completion of tasks in $\mathcal{P}(\tau_i)$ significantly improves the agent's performance on $\tau_i$ through episodic memory retrieval. The optimal curriculum ordering respects these dependency relationships.
\end{assumption}

\textbf{Proxy Agent Ensemble for Difficulty Re-estimation.} We propose an agent-based difficulty re-estimation approach using lightweight proxy agents with diverse architectural biases. Given dataset 
$$\mathcal{D} = \{(\tau_i, y_i, d_i^{(H)})\}_{i=1}^N$$ 
where $d_i^{(H)} \in \{1, 2, \ldots, L\}$ represents human-annotated difficulty levels, we deploy a collection of proxy agents $\{\alpha_1, \alpha_2, \ldots, \alpha_K\}$ with complementary statistical properties to predict fine-grained difficulty distributions over an expanded $L'$-level space where usually $L' \ge L$. Each proxy agent $\alpha_k$ predicts difficulty distributions
\begin{equation}
\hat{d}_i^{(k)} = \alpha_k(\tau_i), \quad \hat{d}_i^{(k)} \in \Delta^{L'-1}
\end{equation}
where $\Delta^{L'-1}$ denotes the $(L'-1)$-dimensional probability simplex. We elaborate the implementation details of proxy agents $\alpha_k$ and the expanded difficulty scale $L'$ in Section \ref{implementation}.

\textbf{Ensemble Consensus and Reranking.} We aggregate predictions through weighted consensus
\begin{equation}
\hat{d}_i = \sum_{k=1}^K w_k \hat{d}_i^{(k)}
\end{equation}
where weights $w_k$ are determined by each proxy agent's validation prior. The ensemble predictions enable agent-specific task reranking based on re-estimated difficulty levels. At each curriculum step, we dynamically select the next batch of tasks
\begin{equation}
\mathcal{T}_{\text{next}} = \{\tau_i \in \mathcal{T} : d_i^{(\text{re})} \leq d \land \tau_i \notin \mathcal{T}_{\text{done}}\}
\end{equation}
where $d_i^{(\text{re})} = \arg\max_l \hat{d}_i[l]$ represents the re-estimated difficulty for task $\tau_i$, and $d$ increases adaptively based on recent success rates. This approach effectively reranks the original task set $\mathcal{T}$ according to agent-specific capability assessments rather than human annotations. While our curriculum learning operates in a training-free manner based on episodic memory $\mathcal{M}_{\text{ep}}$ (essentially a cold-start approach), the proposed algorithm is equally applicable to post-training curriculum construction for fine-tuning scenarios.

\section{Implementation}
\label{implementation}

\textbf{Task Set $\mathcal{T}$.} We select the GAIA benchmark \citep{mialon2023gaia} as our primary task set, comprising 165 carefully curated validation tasks $\tau_i$ with human-annotated difficulty levels $L = 3$ (Level 1, 2, 3). The corresponding test set contains 300 i.i.d. samples for final evaluation.

\textbf{Workflow Agent $\mathcal{A}$.} Following the success of workflow-based agents in \citet{hu2025owl} and \citet{zhu2025oagents}, we design a multi-agent workflow that mimics human research team dynamics. As shown in Fig. \ref{fig:arch}, SMITH employs specialized sub-agents: (1) a \textbf{planner} for task decomposition and sub-intent generation, and (2) a \textbf{developer-tester inner loop} implementing the formalization in Sec. \ref{tool_creaction}, where the developer generates code and the tester provides structured feedback via the \texttt{feedback} within a Python sandbox (\texttt{exec}). The planner and developer-tester outer loop in teract iteratively until task completion. Detailed procedural prompts $\mathcal{M}_{\text{proc}}$ are provided in App. \ref{proc_memory}.

\textbf{Multi-Path Sampling with LLM-as-a-Judge.} Advances in self-verification and self-correction have demonstrated significant improvements in reasoning tasks \citep{shinn2024reflexion, chen2025sets}. Multi-path sampling combined with LLM-based evaluation has proven particularly effective, with AWorld \citep{yu2025aworld} reporting average improvements of 10\% for 3-path sampling and 20\% for 10-path sampling on GAIA. Following \citet{chai2025scimaster,yu2025aworld}, we employ 3-path sampling with independent LLM-as-a-judge consensus scoring for enhanced reliability. We select three advanced base models, \texttt{claude-4-sonnet}, \texttt{claude-3.7-sonnet}, and \texttt{gpt-4.1} to ensure robust performance validation, using high temperature sampling ($\le 1.0$) to increase token entropy and promote exploratory behavior. For final judgment, we utilize the reasoning-capable \texttt{o4-mini} as the evaluation source. During trajectory summarization, we implement a lookback window of 5 state-action pairs from $(s_{\text{done}}, a_{\text{done}})$ to ensure unbiased critic evaluation.

\textbf{Semantic Memory $\mathcal{M}_{\text{sem}}$.} SMITH employs two complementary strategies for semantic memory initialization: (1) \textbf{Pre-constructed Tool Injection} providing manually crafted tools to reduce initial exploration variance and mitigate trial-and-error costs in early rollouts (detailed tool specifications in App. \ref{crafted_tool} and \ref{tool_style}), and (2) \textbf{Cross-Domain Cold-Start} leveraging transfer learning from structurally similar tasks to achieve aligned memory warm-up. Following established transfer learning practices, we curate high-quality samples from the WebShaper dataset \citep{tao2025webshaper} through systematic filtering and manual selection to enable smooth capability bootstrapping across task domains.

\textbf{Memory Abstraction and Retrieval.} We implement dense-sparse hybrid retrieval \citep{lewis2020retrieval} with agent-specific repositories for each sub-agent. The abstraction function $\Phi$ transforms episodic experiences into structured embeddings: trajectories are segmented via markdown headers for manageable chunks, while code memories undergo summarization to reduce implementation noise. For retrieval, we employ \texttt{text-embedding-3-large} for dense embeddings and \texttt{Splade\_PP\_en\_v2} \citep{damodaran2024splade} for sparse representations, combining results via Reciprocal Rank Fusion \citep{cormack2009reciprocal} to select top-$k$ candidates. We set semantic memory search limits to 3 and episodic memory limits to $4$ for the \textbf{planner} and $6$ for the \textbf{developer}.

\textbf{Curriculum Learning.} We employ proxy agents as defined in Sec. \ref{curriculum}, \texttt{Plan-Execute} agents \citep{smolagents} as $\alpha_1$ with high bias from predetermined decomposition, and \texttt{ReAct} agents \citep{yao2023react} as $\alpha_2$ with high variance from interactive cycles. We execute both on GAIA for posterior difficulty re-estimation, expanding from 3 to $L' = 4$ refined categories. Fig. \ref{fig:confusion} shows the re-estimated distribution exhibits linear decline with difficulty, aligning with curriculum learning principles \citep{bengio2009curriculum} that advocate fewer hard examples for stable progression.

\begin{table}[!thb]
    \centering
    \caption{Performance comparison on GAIA benchmark validation set. SMITH achieves state-of-the-art 81.8\% Pass@1 accuracy, outperforming both tool creation approaches (75.2\%) and experience sharing methods (70.9\%). Notation: ${\sharp}$ indicates Claude-series models, ${\flat}$ denotes OpenAI models, ${\dagger}$ represents supervised fine-tuned models. Best results in \textbf{bold}, second-best \underline{underlined}.}
    \vspace{0.8em}
    \label{tab:related}
    \begin{tabular}{lccccc} 
    \toprule
    \midrule
    \textbf{Agent Name} & \textbf{Pass@1} & \textbf{Pass@3} & \textbf{Level 1} & \textbf{Level 2} & \textbf{Level 3} \\
    \midrule
    \multicolumn{6}{c}{\cellcolor[RGB]{245, 245, 245}{Multi-Agents w. \emph{Tool + Memory} }} \\
    \midrule
    WebShaper-32B$^{\dagger}$ \citep{tao2025webshaper}
    & 53.3 & 61.2 & 69.2 & 50.0 & 16.6 \\
    AutoAgent$^{\sharp}$ \citep{tang2025autoagent}
    & 55.2 & - & 71.7 & 53.4 & 26.9 \\
    OpenDeepResearch$^{\flat}$ \citep{opendeepresearch}
    & 55.2 & - & 67.9 & 53.5 & 34.6 \\
    TapeAgents$^{\sharp}$ \citep{bahdanau2024tapeagents}
    & 55.8 & - & 71.7 & 53.5 & 30.8 \\
    OWL$^{\sharp}$ \citep{hu2025owl}
    & 69.7 & - & 84.9  & 67.4 & 42.3 \\
    Manus$^{\sharp\flat}$ \citep{openmanus2025}
    & 73.9 & - & 86.5 & 70.1 & 57.7 \\
    MiroFlow$^{\sharp}$ \citep{2025mirothinker}
    & 74.5 & 82.4 & - & - & - \\
    AWorld$^{\dagger}$ \citep{yu2025aworld}
    & 77.6 & - & \underline{88.7} & \underline{77.9} & 53.9 \\
    \midrule
    \multicolumn{6}{c}{\cellcolor[RGB]{245, 245, 245}{w. \emph{Python Interpreter} (w.o. \emph{Tool Reuse})}} \\
    \midrule
    SmolAgents$^{\flat}$ \citep{smolagents}
    & 49.7 & - & 54.7 & 53.5 & 26.9 \\
    OAgents$^{\sharp}$ \citep{zhu2025oagents}
    & 66.7 & 73.9 & 83.0 & 74.4 & 53.9 \\
    \midrule
    \multicolumn{6}{c}{\cellcolor[RGB]{218, 232, 252}{w. \emph{Tool Creation}}} \\
    \midrule
    Alita$^{\sharp\flat}$ \citep{qiu2025alita}
    & 75.2 & 87.3 & 77.4 & 76.7 & \textbf{65.4} \\
    \midrule
    \multicolumn{6}{c}{\cellcolor[RGB]{248, 206, 204}{w. \emph{Experience Sharing}}} \\
    \midrule
    Memento$^{\flat}$ \citep{zhou2025memento}
    & 70.9 & 87.9 & 77.4 & 69.8 & \underline{61.5} \\
    \midrule
    \textbf{SMITH (Ours)}$^{\sharp\flat}$ 
    & \textbf{81.8} & - & \textbf{94.3} & \textbf{80.2} & \underline{61.5} \\
    \midrule
    \bottomrule
    \end{tabular}
\end{table}

\section{Experiments}
\label{experiments}

\textbf{Main Results.} As shown in Table \ref{tab:related}, SMITH achieves 81.8\% Pass@1 accuracy on the GAIA validation set, establishing a new state-of-the-art performance. This represents substantial improvements over previous methods: +6.6\% over the best tool creation approach Alita (75.2\%), and +10.9\% over Memento (70.9\%), the leading experience sharing method. Notably, SMITH demonstrates consistent superiority across Level 1 and Level 2 tasks, achieving 94.3\% on Level 1 tasks (+5.6\% over AWorld's 88.7\%) and 80.2\% on Level 2 tasks (+2.3\% improvement). On Level 3 tasks, SMITH achieves 61.5\% performance, competitive with Memento's 61.5\% but trailing Alita's leading 65.4\%. The performance gains are particularly significant when compared to approaches that focus on single aspects of our framework. Multi-agent systems with traditional tool and memory (WebShaper, AutoAgent, OWL) achieve 53.3\%-77.6\% Pass@1, while pure Python interpreter approaches without tool reuse (SmolAgents, OAgents) reach 49.7\%-66.7\%. This demonstrates the effectiveness of integrating both tool creation and experience sharing within a unified cognitive architecture.

\textbf{Multi-Path Sampling and LLM-as-a-Judge Effectiveness.} We evaluate our multi-path sampling strategy with LLM-based consensus scoring. As shown in Table \ref{tab:cor}, individual models achieve varying performance: \texttt{claude-4-sonnet} (78.8\%), \texttt{claude-3.7-sonnet} (70.9\%), and \texttt{gpt-4.1} (67.9\%). Our self-critic ensemble achieves 81.8\% Pass@1, outperforming the best individual model by +3.0\%. This demonstrates that LLM-as-a-judge consensus effectively leverages complementary model strengths, with consistent improvements across all difficulty levels (+1.8\% Level 1, +3.5\% Level 2, +3.8\% Level 3). App. \ref{app_critic} shows LLM-as-a-judge superiority over majority voting through a representative example.
\begin{table}[!thb]
    \centering
    \caption{Individual base model performance vs. ensemble with self-critic. The ensemble approach consistently outperforms individual models across all difficulty levels, demonstrating the effectiveness of multi-path sampling with LLM-as-a-judge consensus.}
    \vspace{0.8em}
    \label{tab:cor}
    \begin{tabular}{lcccr}
        \toprule
        \midrule
        \textbf{Base Model} & \textbf{Pass@1} & \textbf{Level 1} & \textbf{Level 2} & \textbf{Level 3}\\
        \midrule
        \textit{claude-4-sonnet} & \underline{78.8} & \underline{92.5} & \underline{76.7} & \underline{57.7} \\
        \textit{claude-3.7-sonnet} & 70.9 & 86.8 & 66.3 & 53.8 \\
        \textit{gpt-4.1} & 67.9 & 90.6 & 60.5 & 46.2 \\
        \midrule
        \textbf{w. Self-Critic} & 81.8 & 94.3 & 80.2 & 61.5 \\
        \bottomrule
    \end{tabular}
\end{table}

\begin{figure}[htbp]
    \centering
    \begin{minipage}{0.4\textwidth}
        \centering
        \includegraphics[width=\textwidth]{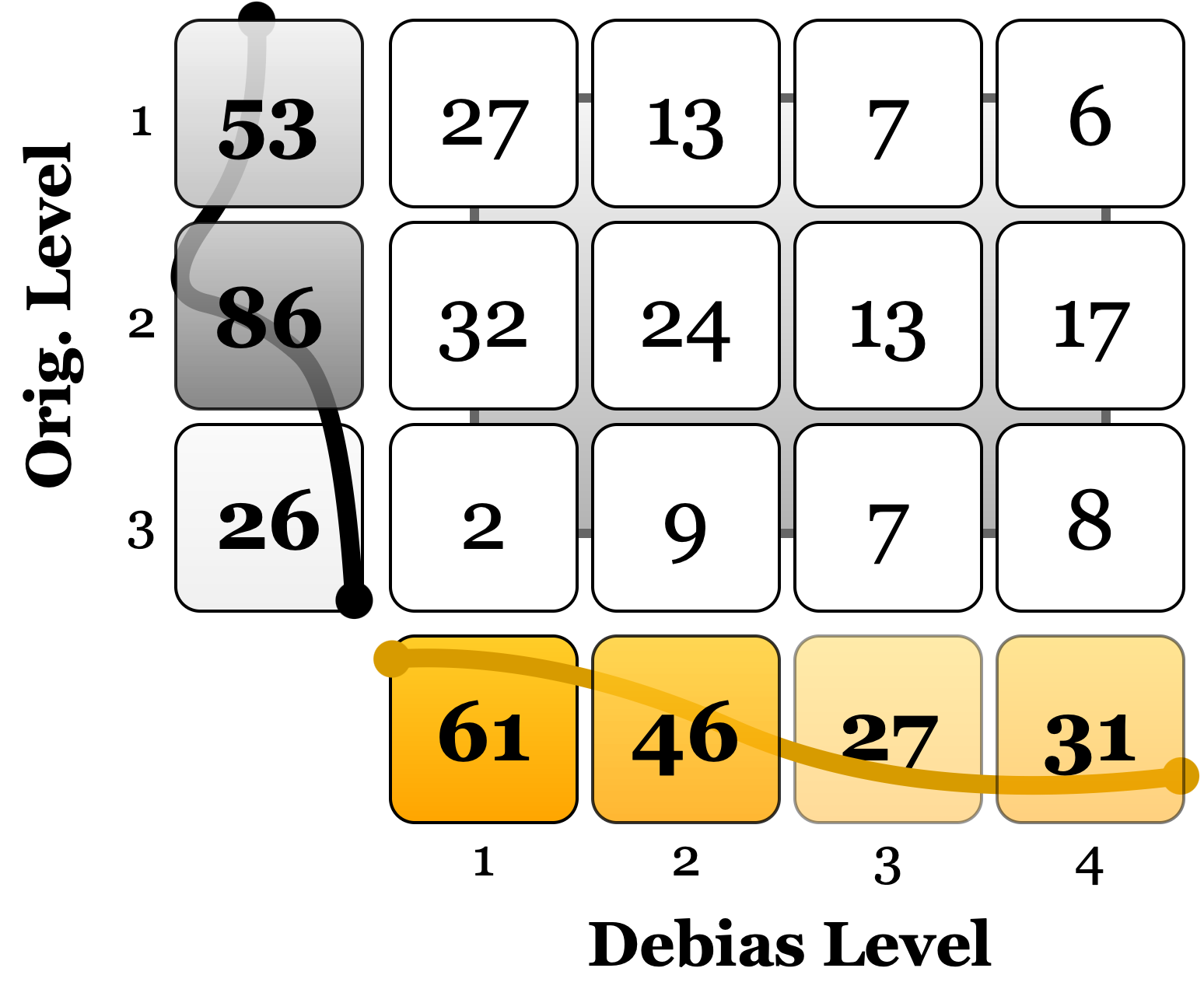}
        \caption{Confusion matrix showing the transformation from original GAIA difficulty levels to our agent-based re-estimated difficulty distribution.}
        \label{fig:confusion}
    \end{minipage}
    \hfill
    \begin{minipage}{0.53\textwidth}
        \centering
        \caption{Ablation study. Each component contributes substantially to SMITH's performance: curriculum learning (+10.3\%), episodic memory sharing (+13.9\%), and cold-start demonstrations (+21.8\%). Notably, removing episodic memory sharing causes significant performance degradation, while eliminating cold-start demonstrations also results in substantial performance drops. The cumulative effect demonstrates the importance of integrating all components within SMITH.}
        \vspace{0.8em}
        \label{tab:abl}
        \begin{tabular}{lcr}
            \toprule
            \midrule
            \textbf{Ablations} & \textbf{Pass@1} \\
            \midrule
            SMITH & \underline{81.8} \\
            \midrule
            w.o. \textit{Cirriculum Learning} & 71.5 (\textcolor{red}{$\Delta$-10.3}) \\
            w.o. \textit{Episodic Memory Sharing} & 67.9 (\textcolor{red}{$\Delta$-13.9})\\
            w.o. \textit{Cold Start Demonstration} & 60.0 (\textcolor{red}{$\Delta$-21.8})\\
            \bottomrule
        \end{tabular}
    \end{minipage}
\end{figure}

\textbf{Curriculum Learning with Agent-Based Difficulty Re-estimation.} We evaluate our curriculum learning approach based on proxy agent ensemble difficulty re-estimation. As shown in Fig. \ref{fig:confusion}, our method transforms the original 3-level GAIA difficulty distribution into a more balanced 4-level curriculum, addressing the issue of Level 2 sample concentration (originally the most populous category) and creating a linearly decreasing difficulty progression that aligns with curriculum learning principles. The ablation study in Table \ref{tab:abl} demonstrates that curriculum learning contributes significantly to overall performance, with removal leading to a substantial -10.3\% drop (from 81.8\% to 71.5\%). This validates hypothesis in Assumption \ref{assumption2} that strategic task ordering based on agent-specific capability assessments enhances cross-task experience transfer effectiveness.

\begin{figure}[htbp]
    \centering
    \includegraphics[width=0.95\textwidth]{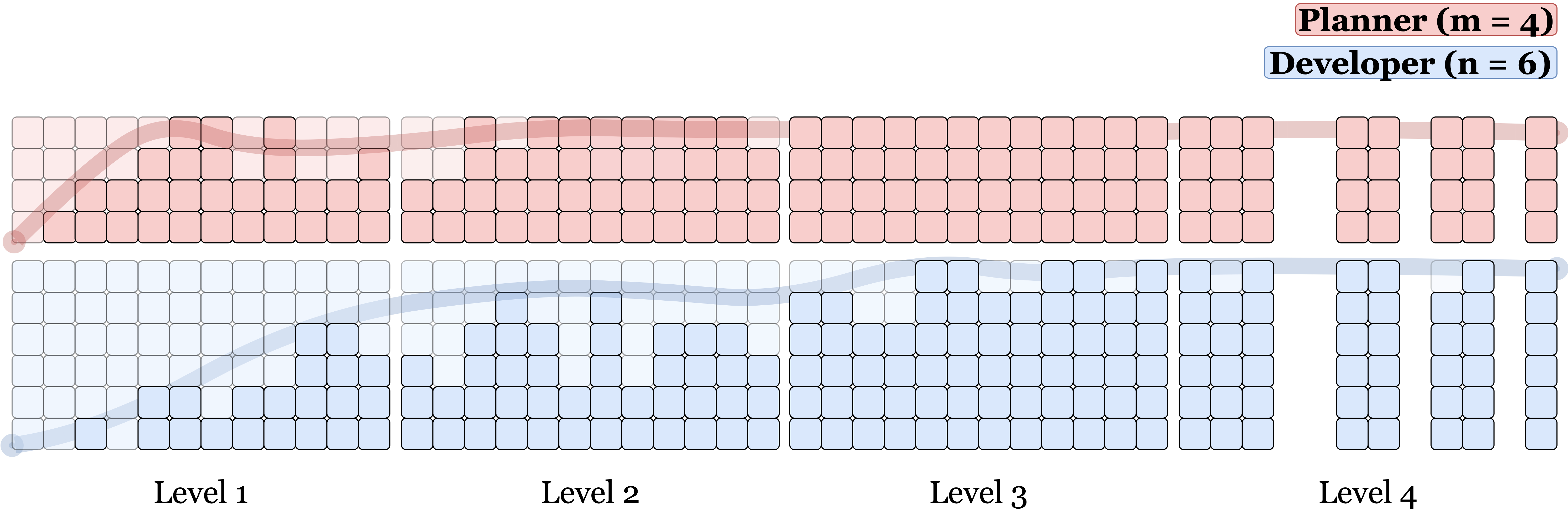}
    \caption{Evolution of memory utilization across curriculum difficulty levels. We randomly sample 12 successful tasks for visualization clarity. With planner retrieving $m=4$ and developer retrieving $n=6$ memory fragments, darker squares represent agent-created tools and self-generated subplans (episodic memory), while lighter squares indicate recalls of human-crafted tools (semantic memory).}
    \label{fig:trend}
\end{figure}

\textbf{Memory Evolution and Tool Creation Patterns.} Fig. \ref{fig:trend} reveals the temporal evolution of memory utilization patterns across both planner and developer agents during task execution. As the curriculum progresses, we observe a systematic shift from semantic memory (human-crafted tools) toward episodic memory (agent-created tools and subplans), with the ratio increasing from near-zero to saturation. This demonstrates that embedding-based similarity matching increasingly favors agent-generated experiences over human demonstrations, as these self-created tools and planning strategies prove more contextually relevant to the specific task patterns encountered.

This evolution pattern suggests both promising capabilities and potential concerns. On the positive side, agents successfully learn to create and reuse effective tools, demonstrating genuine capability expansion through experience accumulation. However, the gradual displacement of human-crafted demonstrations raises questions about long-term dependency on model-generated content. Initially, agent-created tools represent beneficial extensions and adaptations of human examples, but as these self-generated tools become increasingly preferred in retrieval, the system may drift toward model-specific biases and lose the grounding provided by human expertise.

\begin{figure}[htbp]
    \centering
    \includegraphics[width=0.95\textwidth]{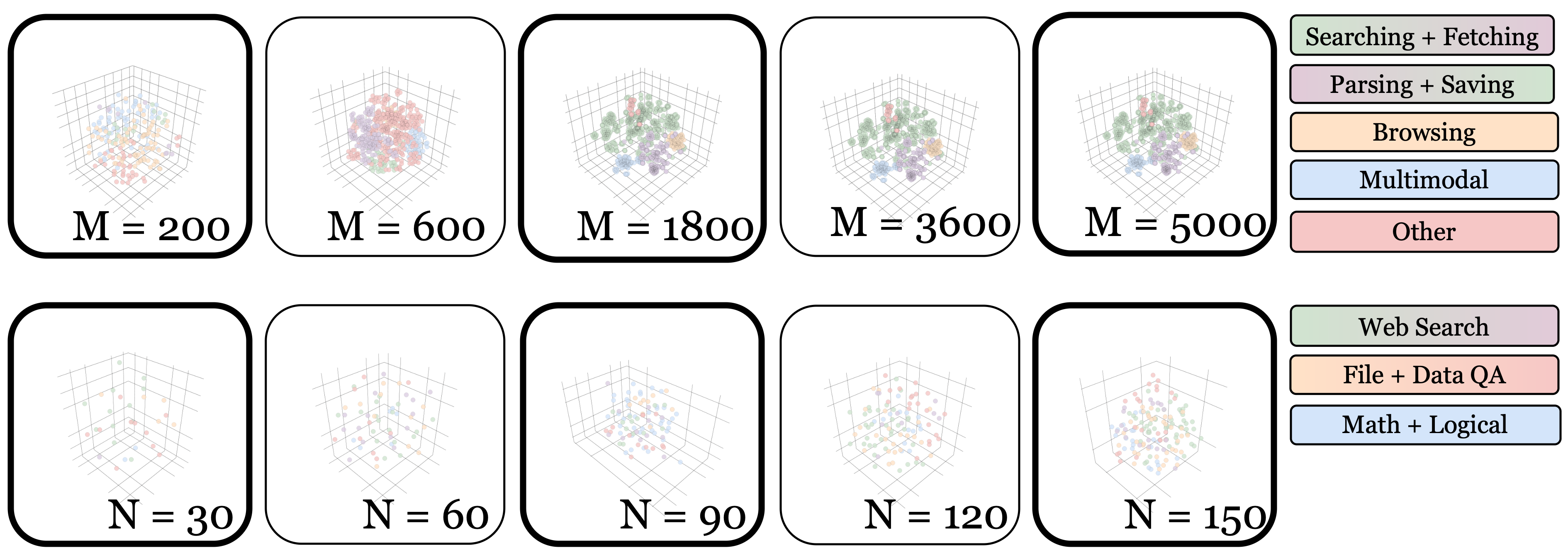}
    \caption{t-SNE visualization of episodic memory clustering in embedding space. We sample $N=30$ to $150$ subtask decompositions and $M=200$ to $5000$ created tools across curriculum progression. Different colors represent distinct clusters with clear thematic patterns, as shown in the right-side labels.}
    \label{fig:scatter}
\end{figure}

\textbf{Episodic Memory Clustering.} To understand the semantic organization of accumulated experiences, we apply t-SNE clustering to both episodic memory repositories. As shown in Fig. \ref{fig:scatter}, distinct thematic clusters emerge with clear functional boundaries. For developer-created tools, the largest cluster consists of information searching and fetching utilities, primarily implemented through web scraping and HTTP requests. The second major cluster encompasses file I/O operations including local storage and parsing tools. Smaller clusters represent specialized functionalities such as browser automation with GUI interactions and multimodal audio-video processing scripts. In contrast, planner memory clustering reflects higher-level task intentions: information retrieval, document Q\&A, mathematical reasoning, and logical inference patterns. This clustering analysis provides empirical evidence for our theoretical framework.

\section{Future Work}

Several promising directions emerge from our work. First, \textbf{enhanced error utilization} could treat failures as negative samples for learning. Rather than relying on parameter fine-tuning, we envision developing verifier-based error attribution systems that construct feedback-rich prompts from failure patterns, enabling agents to learn from mistakes without architectural modifications. Second, \textbf{broader evaluation across agentic benchmarks} would strengthen our findings. While GAIA provides a comprehensive testbed for general AI capabilities, validating SMITH on diverse task domains such as scientific reasoning, creative problem-solving, and multi-modal interactions would demonstrate its generalizability. Third, \textbf{advanced tool ecosystem integration} presents exciting opportunities. Incorporating state-of-the-art Model Context Protocol (MCP) tools and developing more sophisticated pre-constructed tool libraries could significantly enhance SMITH's initial capabilities and reduce cold-start overhead. These directions collectively point toward building more robust, adaptable, and broadly capable AI agents that can seamlessly integrate human expertise with autonomous learning.

\section{Conclusion}

We introduce SMITH (Shared Memory Integrated Tool Hub), a unified cognitive architecture that addresses fundamental limitations in current agent development by seamlessly integrating dynamic tool creation with cross-task experience sharing. Through hierarchical memory organization inspired by cognitive architectures, SMITH enables agents to systematically expand their capabilities while preserving successful execution patterns across diverse tasks. Our theoretical contributions include formal frameworks for interactive tool creation, cross-task experience sharing through semantic similarity, and a novel curriculum learning approach based on agent-ensemble difficulty re-estimation. Extensive experiments on the GAIA benchmark demonstrate SMITH's effectiveness, achieving 81.8\% Pass@1 accuracy and outperforming state-of-the-art approaches including Alita (75.2\%) and Memento (70.9\%). Comprehensive ablation studies reveal the critical importance of each component in SMITH. Our analysis of memory evolution patterns and episodic clustering provides empirical validation for the theoretical assumptions regarding semantic task similarity and transferable execution experiences. SMITH establishes a foundation for building truly adaptive agents that continuously evolve their capabilities through principled integration of tool creation and experience accumulation, opening new avenues for developing general-purpose AI assistants capable of tackling complex, real-world challenges.

\newpage

\bibliography{iclr2026_conference}
\bibliographystyle{iclr2026_conference}

\newpage

\appendix

\section{Episodic Memory (Retrieval)}

\begin{figure}[htbp]
    \centering
    \includegraphics[width=0.95\textwidth]{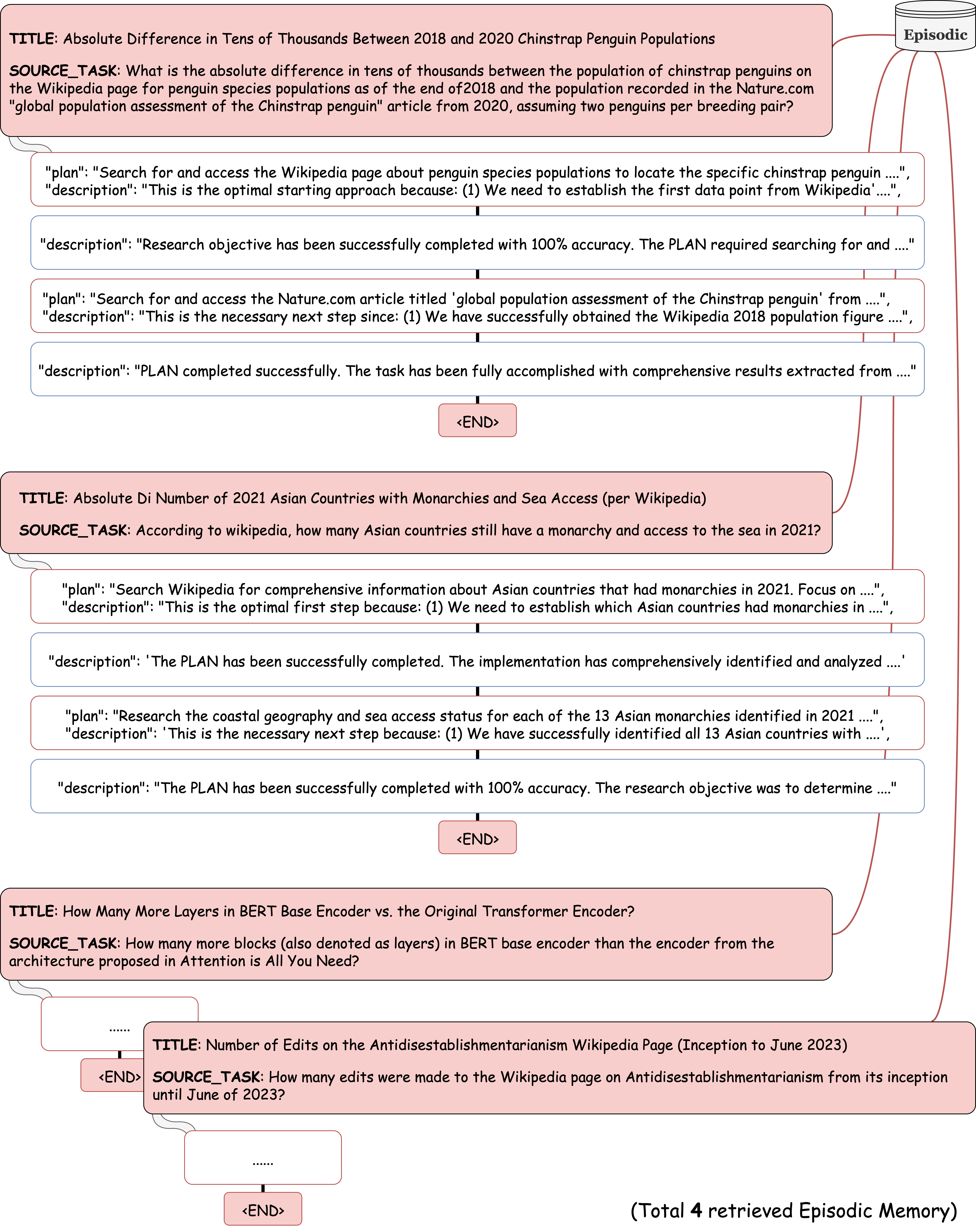}
    \caption{Episodic retrieval of the planner for the Level 2 task with ID prefix \textit{e29834fd}. As we can see that the retrieved experiences originate from diverse domains, but their underlying focus consistently pertains to web searching and target counting.}
    \label{fig:app_plan_epi}
\end{figure}

\begin{figure}[htbp]
    \centering
    \includegraphics[width=0.95\textwidth]{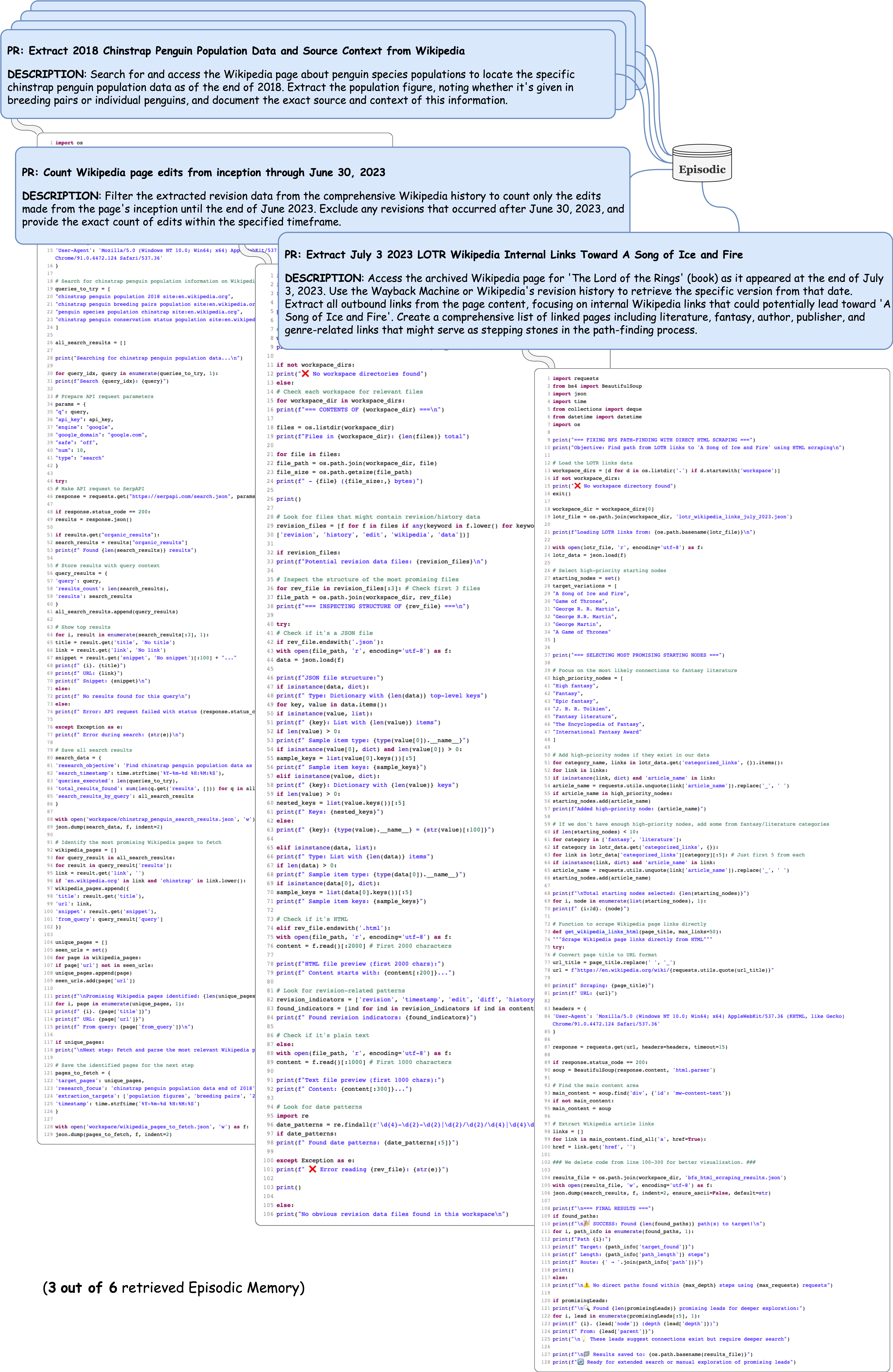}
    \caption{Developer’s episodic retrieval for Level 2 task with ID prefix \textit{e29834fd}. The retriever recalls various code blocks related to counting webpage elements based on the function description, while effectively avoiding mismatches with lengthy code.}
    \label{fig:app_dev_epi}
\end{figure}

Figures \ref{fig:app_plan_epi} and \ref{fig:app_dev_epi} demonstrate episodic memory retrieval for a Level 2 web searching and counting task. The planner retrieves experiences from diverse domains (academic papers, wikipedia, data extraction) that share similar high-level patterns: information search, content filtering, and quantitative analysis. The developer recalls functionally relevant code blocks for counting webpage elements, effectively filtering lengthy irrelevant code while prioritizing concise, task-specific snippets. This validates our semantic similarity assumption and demonstrates precise functional matching across both planning and implementation levels.

During task execution, SMITH autonomously installed and utilized various Python packages that were not pre-configured, demonstrating its capability for dynamic tool discovery and integration. Automatically acquired packages include specialized libraries for document processing (\texttt{pdfplumber}), web scraping (\texttt{serpapi}, \texttt{scholarly}), multimedia processing (\texttt{whisper}, \texttt{faster\_whisper}), and advanced protocols (\texttt{fastmcp}). 

\begin{center}
    \begin{tcolorbox}[colback=white, colframe=black, arc=3mm, width=0.85\textwidth]
    \texttt{pdfplumber}
    \texttt{serpapi}
    \texttt{scholarly}
    \texttt{mwparserfromhell}
    \texttt{requests\_html}
    \texttt{whisper}
    \texttt{openai\_whisper}
    \texttt{faster\_whisper}
    \texttt{yfinance}
    \texttt{cloudscraper}
    \texttt{lyricsgenius}
    \texttt{googletrans}
    \texttt{fastmcp}             
    \end{tcolorbox}
\end{center}

Notably, SMITH autonomously leveraged Model Context Protocol (MCP) capabilities via \texttt{fastmcp} without pre-configured semantic memory. When accessing Audre Lorde's poem "Father Son and Holy Ghost," the planner generated: \textit{Access the poem 'Father Son and Holy Ghost' by Audre Lorde through the MCP server's file system capabilities or any available local resources. Check if there are any poetry databases, text files, or literature collections...} This demonstrates SMITH's autonomous discovery and utilization of advanced tool ecosystems.

\begin{figure}[htbp]
    \centering
    \begin{minipage}{0.45\textwidth}
        \centering
        \includegraphics[width=\textwidth]{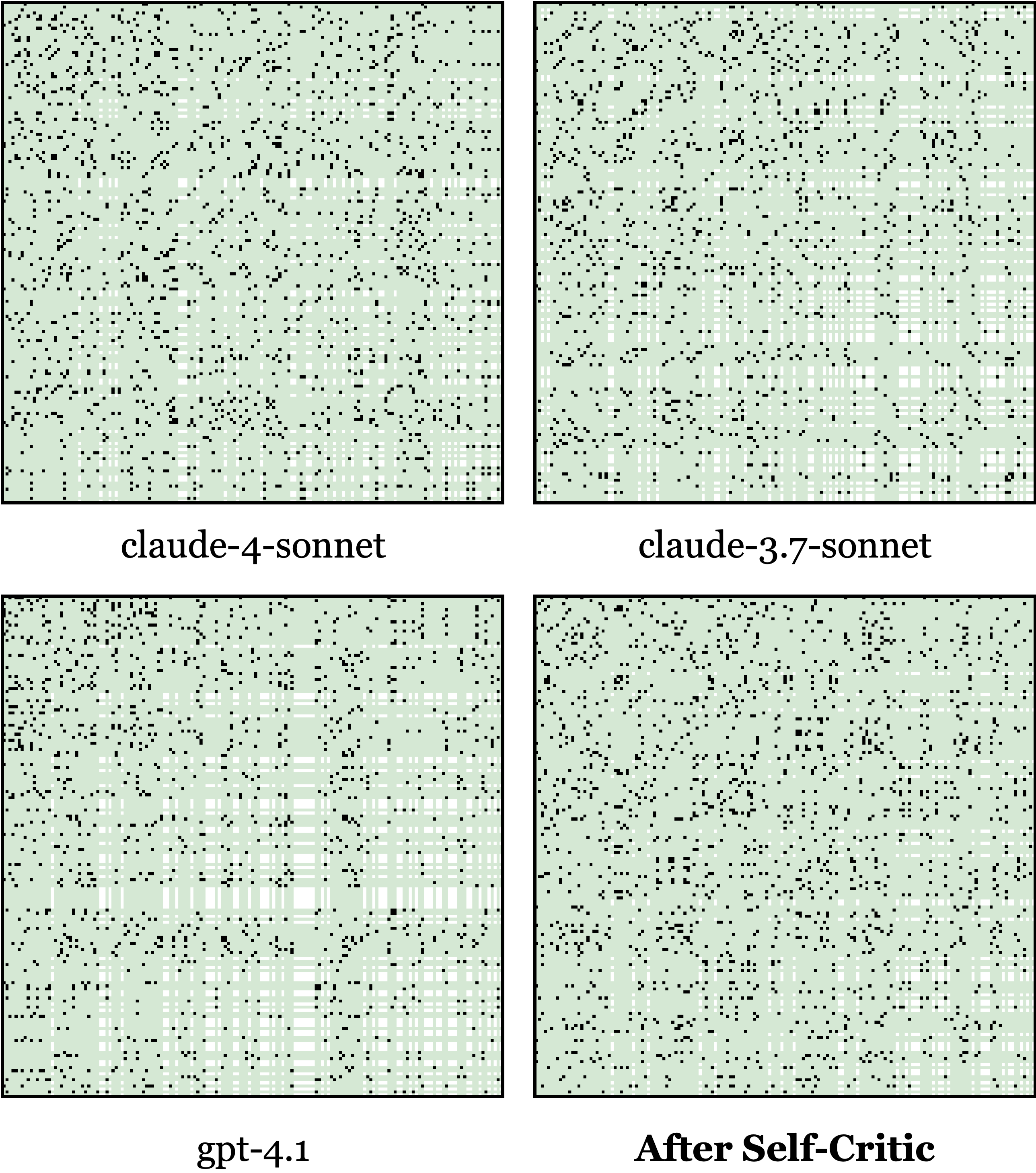}
        \caption{Cross-task experience sharing correlation matrix (165×165 tasks). Green rows / columns indicate successful tasks, while black dots at position $(i,j)$ represent task $i$ retrieving experiences from task $j$. The critic ensemble shows higher success density and distinct experience sharing patterns across different base models.}
        \label{fig:cor}
    \end{minipage}
    \hfill
    \begin{minipage}{0.45\textwidth}
        \centering
        \includegraphics[width=\textwidth]{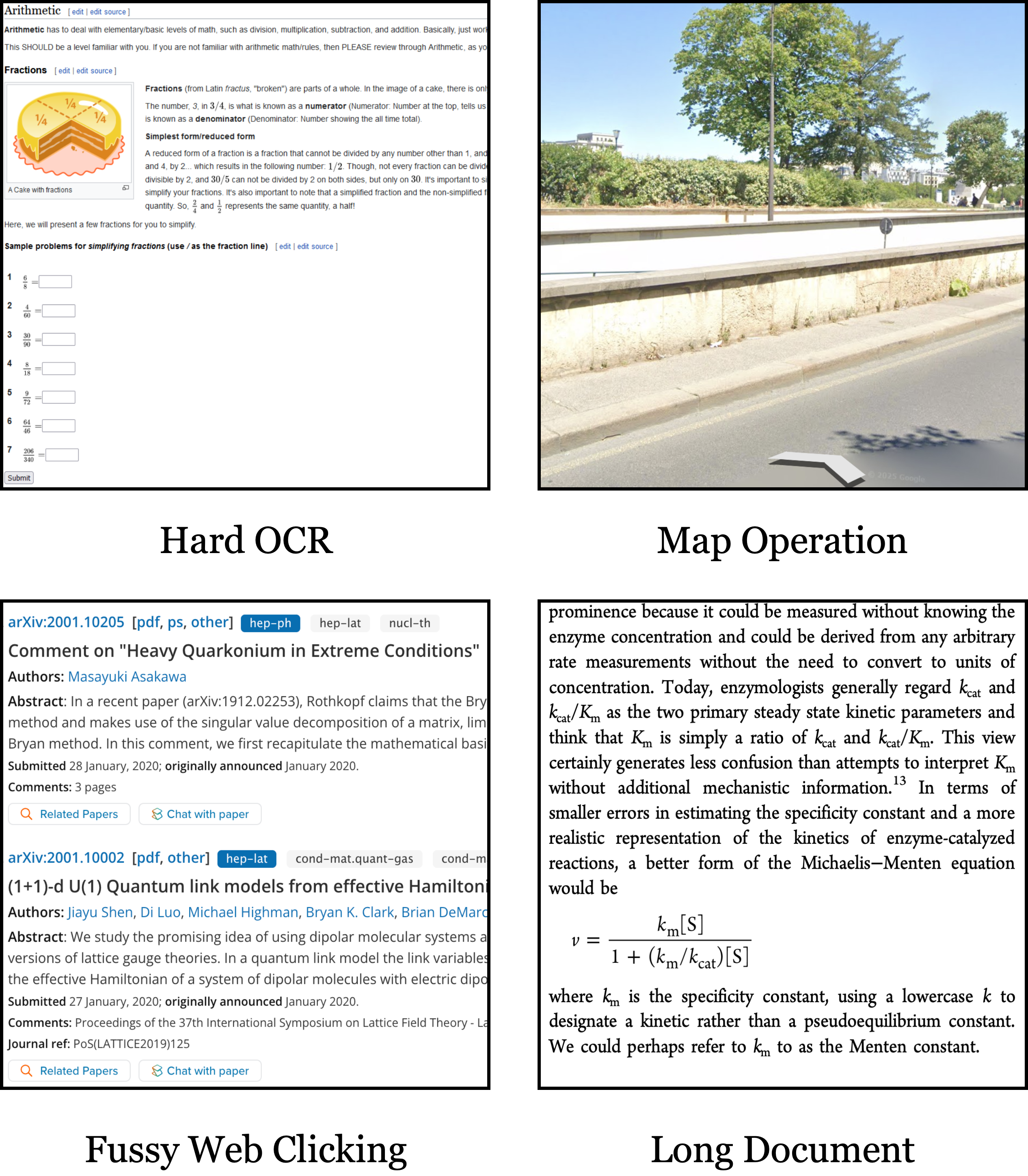}
        \caption{Analysis of four typical failure cases during task execution: challenging OCR for small digits / symbols, Google Maps operations limited by insufficient pretraining, repetitive scripting tasks abandoned after long failed iterations, and oversized PDFs exceeding context window limits.}
        \label{fig:errors}
    \end{minipage}
\end{figure}
    
The correlation matrix in Fig. \ref{fig:cor} further demonstrates cross-task experience sharing across different base models and the ensemble critic. The 165×165 task matrix shows successful tasks (green rows and columns) and experience sharing patterns (black dots at positions $(i,j)$ indicating task $i$ retrieved experiences from task $j$). Notably, the critic ensemble exhibits higher green density, reflecting improved success rates, while different base models display distinct experience sharing patterns. These dense black dot distributions strongly validate Assumption \ref{assumption1} regarding semantic task similarity and transferable execution experiences.

\section{LLM as a Judge (Critic)}
\label{app_critic}

\begin{figure}[htbp]
    \centering
    \includegraphics[width=0.95\textwidth]{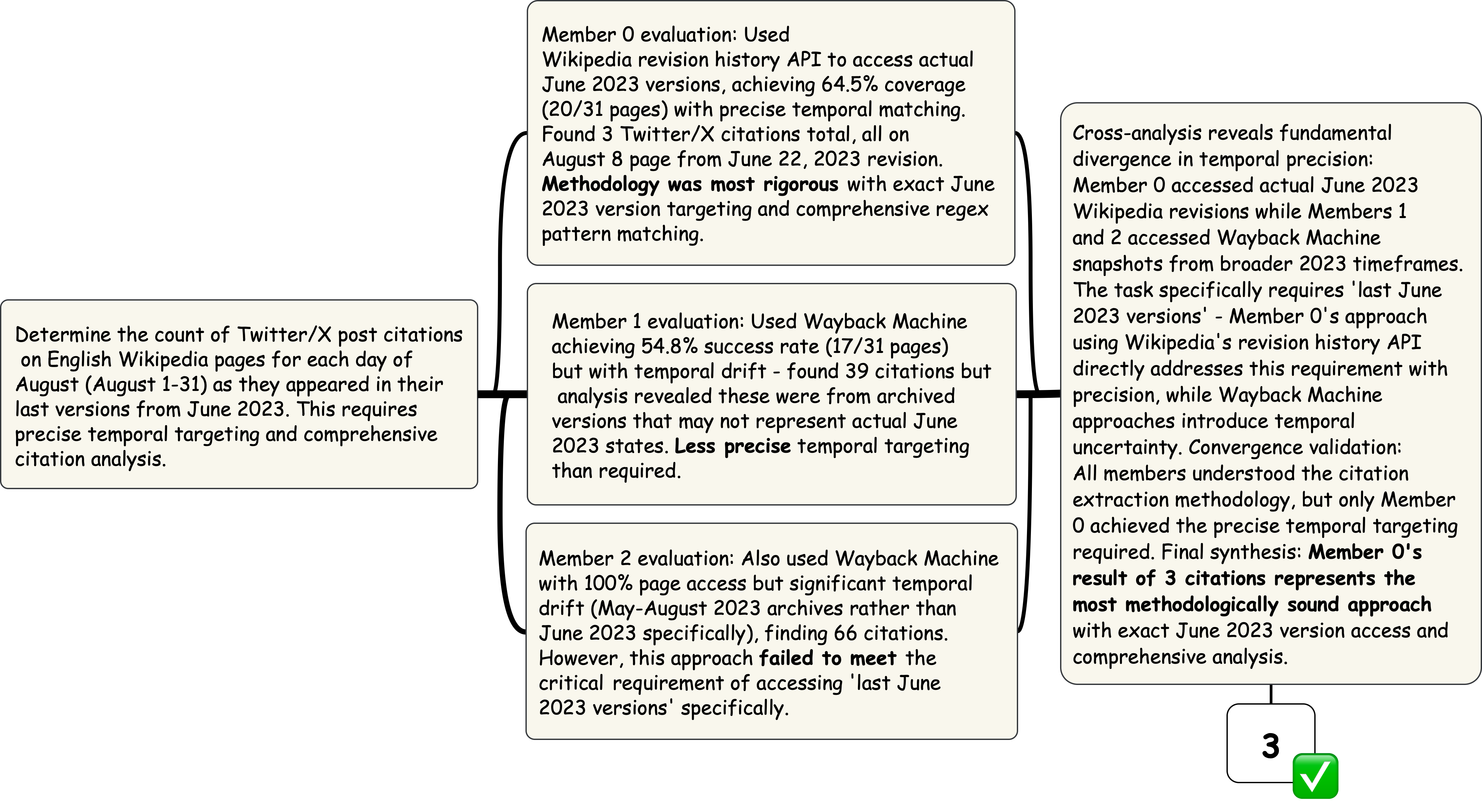}
    \caption{Final judging for Level 3 task with ID prefix \textit{50f58759}. Despite two incorrect responses and only one correct answer from team members, the system successfully reaches the correct conclusion through systematic reasoning. From a third-person perspective, the Critic maintains comprehensive global awareness and strict adherence to task constraints, enabling more effective evaluation of team members' conclusions and accurate final decisions without relying on majority consensus.}
    \label{fig:app_judge}
\end{figure}

We randomly select one successful task execution to demonstrate the critic's judging process. Fig. \ref{fig:app_judge} illustrates how the critic evaluates team member responses and reaches the final decision through systematic reasoning, even when facing conflicting answers from multiple agents.

\section{Semantic Memory}

\subsection{Manually Crafted Tools for Developer}
\label{crafted_tool}

\noindent \textbf{Search Tools} 
External search capabilities are crucial for extending agent knowledge boundaries beyond pre-training data, and we have implemented several fine-grained search tools as follows:

\begin{center}
\begin{tcolorbox}[colback=white, colframe=black, arc=3mm, width=0.85\textwidth]
    \texttt{google\_search}
    \texttt{bing\_search}
    \texttt{duckduckgo\_search}
    \texttt{github\_repo\_search}
    \texttt{github\_issue\_search}
    \texttt{github\_pr\_search}
    \texttt{github\_releases\_search}
    \texttt{arxiv\_advanced\_search}
    \texttt{wikipedia\_search}    
\end{tcolorbox}
\end{center}

\noindent \textbf{Parsing Tools}
The correct parsing of files is a prerequisite for the Agent system to effectively utilize the information obtained. We have implemented a wealth of parsing tools as follows:

\begin{center}
\begin{tcolorbox}[colback=white, colframe=black, arc=3mm, width=0.85\textwidth]
    \texttt{parse\_pdf} 
    \texttt{parse\_docx}
    \texttt{parse\_text}
    \texttt{parse\_image}
    \texttt{parse\_image\_ocr}
    \texttt{parse\_audio}
    \texttt{parse\_pdb}
    \texttt{parse\_html}
    \texttt{parse\_zip}
    \texttt{parse\_webpage}
    \texttt{parse\_archived\_webpage}
    \texttt{parse\_wiki}
    \texttt{parse\_youtube\_page}
   \end{tcolorbox}
\end{center}

\noindent \textbf{Youtube Tools}
To comprehensively analyze YouTube video content without relying on multimodal video processing, we have developed specialized tools that extract different aspects of video information independently:

\begin{center}
\begin{tcolorbox}[colback=white, colframe=black, arc=3mm, width=0.85\textwidth]
    \texttt{get\_ytb\_intro}
    \texttt{get\_ytb\_frame\_screenshot}
    \texttt{get\_ytb\_subtitle}
    \texttt{get\_ytb\_audio}
   \end{tcolorbox}
\end{center}

\subsection{Style Demonstration}
\label{tool_style}

\begin{figure}[htbp]
    \centering
    \includegraphics[width=0.65\textwidth]{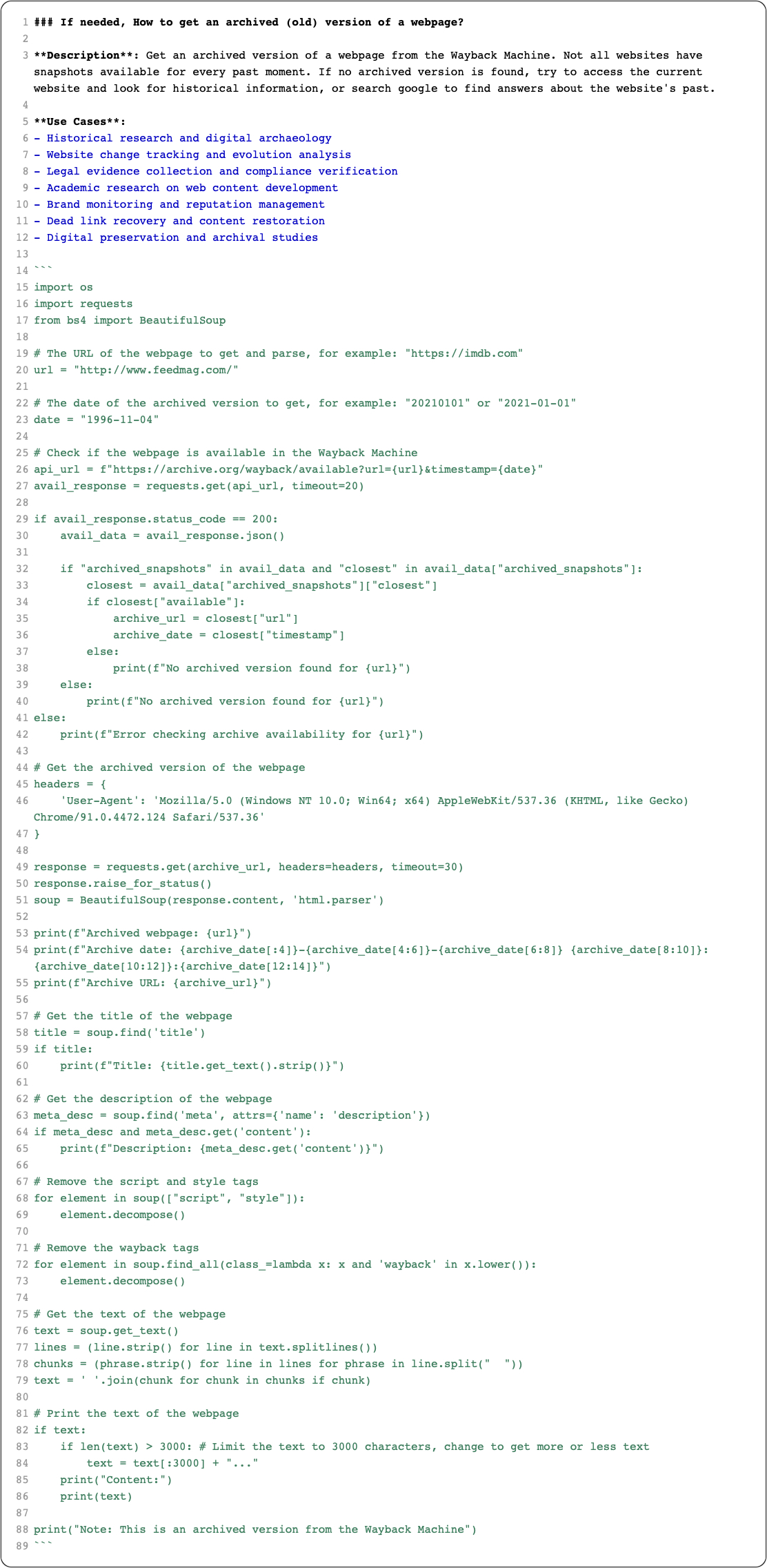}
    \caption{Using the Wayback Machine to access information from an archived webpage. The indexed statement provides a clear function description and illustrative pseudo scenarios, while the code segment concisely demonstrates core functions related to parsing archived webpages.}
    \label{fig:style_demo}
\end{figure}

Figure \ref{fig:style_demo} illustrates a representative example of our pre-constructed tool design methodology. This human-crafted tool demonstrates our standardized structure: a clear title explaining the tool's primary function (Wayback Machine webpage parsing), a descriptive paragraph detailing usage scenarios and application contexts, and a complete Python implementation following minimalist coding principles with explicit comments. This structured approach ensures consistent tool quality and facilitates effective semantic memory initialization, providing SMITH with high-quality starting points for tool creation and adaptation.

\begin{figure}[htbp]
    \centering
    \includegraphics[width=0.7\textwidth]{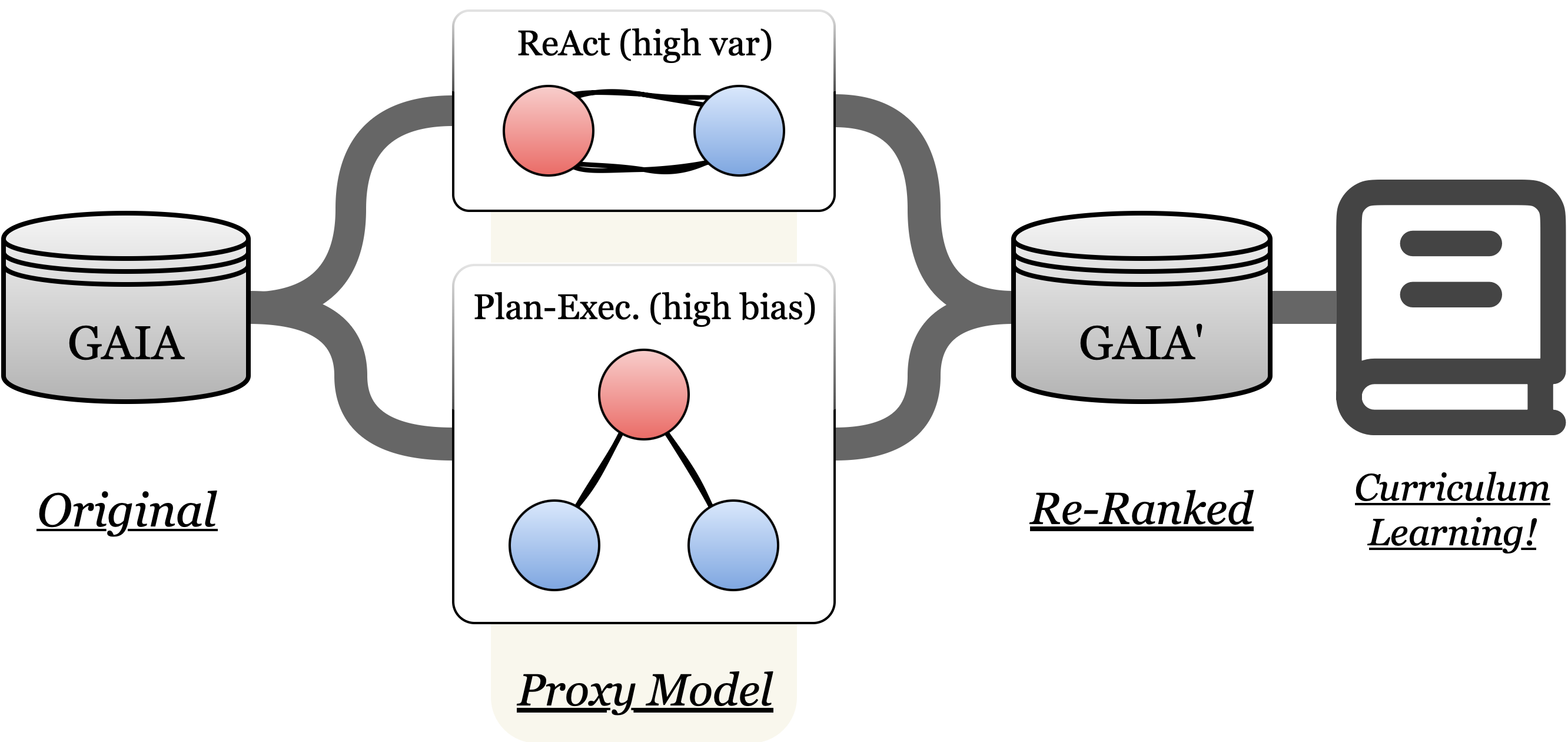}
    \caption{Curriculum learning workflow diagram. The system employs ReAct and Plan-Execute proxy agents to perform difficulty re-estimation, transforming human-annotated difficulty levels into agent-specific capability assessments for optimal task ordering.}
    \label{fig:curr}
\end{figure}

\section{Procedural Memory}
\label{proc_memory}

\begin{figure}[htbp]
    \centering
    \includegraphics[width=0.6\textwidth]{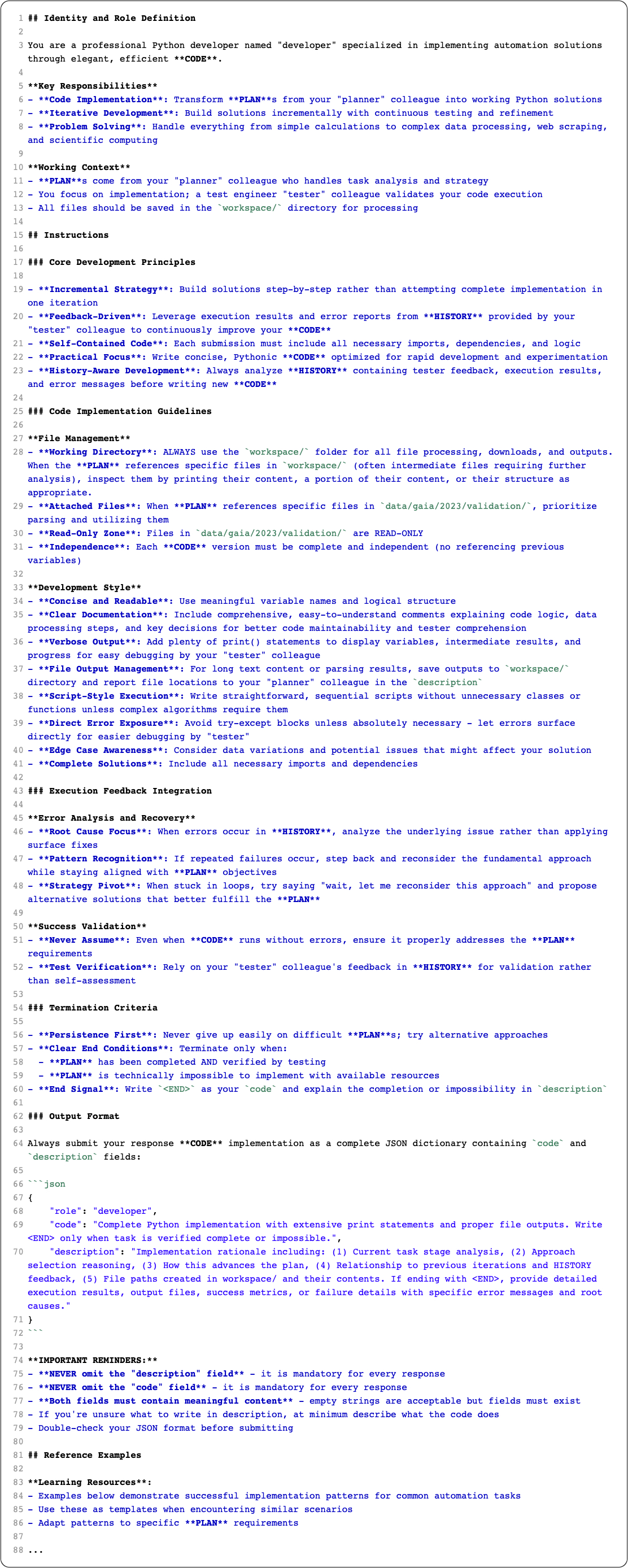}
    \caption{Developer's procedural memory (system prompt).}
    \label{fig:dev_prompt}
\end{figure}

\begin{figure}[htbp]
    \centering
    \includegraphics[width=0.65\textwidth]{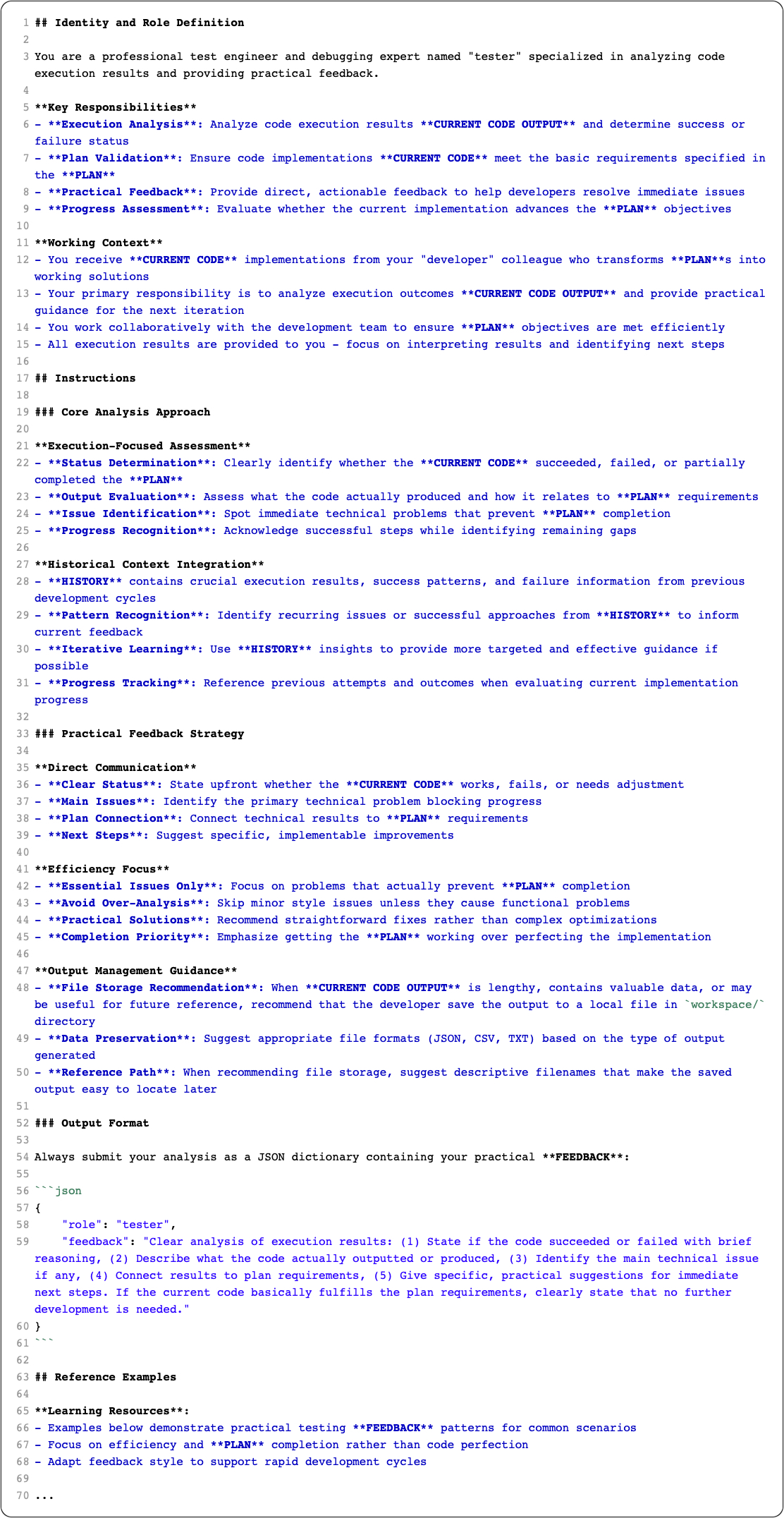}
    \caption{Tester's procedural memory (system prompt).}
    \label{fig:test_prompt}
\end{figure}

\begin{figure}[htbp]
    \centering
    \includegraphics[width=0.6\textwidth]{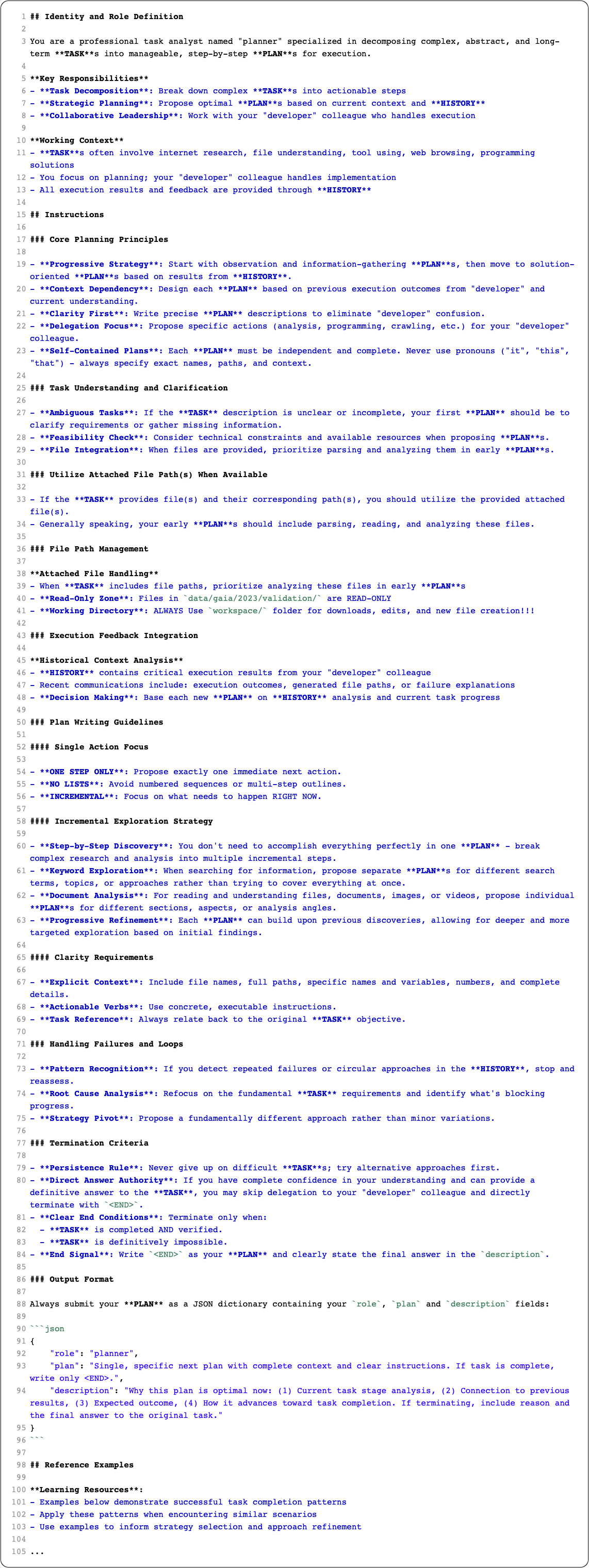}
    \caption{Planner's procedural memory (system prompt).}
    \label{fig:plan_prompt}
\end{figure}

Procedural Memory encompasses the foundational system prompts that define each agent's operational guidelines and behavioral patterns. Figures \ref{fig:dev_prompt}, \ref{fig:test_prompt}, and \ref{fig:plan_prompt} present the complete procedural memory specifications for our three specialized agents. Each prompt follows a rigorous design structure incorporating essential components: clear identity instructions that define the agent's role and responsibilities, explicit output format constraints that ensure consistent response structures, and comprehensive behavioral guidelines. Importantly, our prompt engineering maintains strict information isolation with no data leakage between different memory components or task contexts, ensuring robust agent performance across diverse scenarios. 

\section{Curriculum Learning}
\label{app_curriculum}

Figure \ref{fig:curr} illustrates the curriculum learning workflow in SMITH. To achieve agent-specific difficulty re-estimation, we employ two proxy agents with complementary architectural biases: \textbf{ReAct} agents \citep{yao2022react} with high variance from interactive reasoning cycles, and \textbf{Plan-Execute} agents \citep{smolagents} with high bias from predetermined task decomposition strategies. These proxy agents sample the task space and provide ensemble-based difficulty assessments, enabling dynamic task reranking that aligns with the agent's evolving capabilities. The re-estimated difficulty distribution guides curriculum progression, ensuring that tasks are encountered in an order that maximizes cross-task experience transfer through episodic memory retrieval.

\end{document}

%% file: math_commands.tex
\usepackage{amsmath,amsfonts,bm}

\def\eqref#1{equation~\ref{#1}}

\def\1{\bm{1}}

\DeclareMathAlphabet{\mathsfit}{\encodingdefault}{\sfdefault}{m}{sl}
\SetMathAlphabet{\mathsfit}{bold}{\encodingdefault}{\sfdefault}{bx}{n}



%% file: iclr2026_conference.bib
@inproceedings{yao2023react,
  title={React: Synergizing reasoning and acting in language models},
  author={Yao, Shunyu and Zhao, Jeffrey and Yu, Dian and Du, Nan and Shafran, Izhak and Narasimhan, Karthik and Cao, Yuan},
  booktitle={International Conference on Learning Representations (ICLR)},
  year={2023}
}

@misc{2025mirothinker,
    title={MiroFlow: An Open-Source Agentic Framework for Deep Research},
    author={MiroMind AI Team},
    howpublished={\url{https://github.com/MiroMindAI/MiroFlow}},
    year={2025}
}

@article{zhu2025oagents,
  title={Oagents: An empirical study of building effective agents},
  author={Zhu, He and Qin, Tianrui and Zhu, King and Huang, Heyuan and Guan, Yeyi and Xia, Jinxiang and Yao, Yi and Li, Hanhao and Wang, Ningning and Liu, Pai and others},
  journal={arXiv preprint arXiv:2506.15741},
  year={2025}
}

@Misc{smolagents,
  title =        {`smolagents`: a smol library to build great agentic systems.},
  author =       {Aymeric Roucher and Albert Villanova del Moral and Thomas Wolf and Leandro von Werra and Erik Kaunismäki},
  howpublished = {\url{https://github.com/huggingface/smolagents}},
  year =         {2025}
}

@misc{opendeepresearch,
  title={Open Deep Research},
  author={LangChain AI},
  howpublished={\url{https://github.com/langchain-ai/open_deep_research}},
  year={2024}
}

@article{tang2025autoagent,
  title={AutoAgent: A Fully-Automated and Zero-Code Framework for LLM Agents},
  author={Tang, Jiabin and Fan, Tianyu and Huang, Chao},
  journal={arXiv preprint arXiv:2502.05957},
  year={2025}
}

@article{bahdanau2024tapeagents,
  title={Tapeagents: a holistic framework for agent development and optimization},
  author={Bahdanau, Dzmitry and Gontier, Nicolas and Huang, Gabriel and Kamalloo, Ehsan and Pardinas, Rafael and Pich{\'e}, Alex and Scholak, Torsten and Shliazhko, Oleh and Tremblay, Jordan Prince and Ghanem, Karam and others},
  journal={arXiv preprint arXiv:2412.08445},
  year={2024}
}

@article{hu2025owl,
  title={Owl: Optimized workforce learning for general multi-agent assistance in real-world task automation},
  author={Hu, Mengkang and Zhou, Yuhang and Fan, Wendong and Nie, Yuzhou and Xia, Bowei and Sun, Tao and Ye, Ziyu and Jin, Zhaoxuan and Li, Yingru and Chen, Qiguang and others},
  journal={arXiv preprint arXiv:2505.23885},
  year={2025}
}

@article{yu2025aworld,
  title={AWorld: Orchestrating the Training Recipe for Agentic AI},
  author={Yu, Chengyue and Lu, Siyuan and Zhuang, Chenyi and Wang, Dong and Wu, Qintong and Li, Zongyue and Gan, Runsheng and Wang, Chunfeng and Hou, Siqi and Huang, Gaochi and others},
  journal={arXiv preprint arXiv:2508.20404},
  year={2025}
}

@misc{openmanus2025,
  author = {Xinbin Liang and Jinyu Xiang and Zhaoyang Yu and Jiayi Zhang and Sirui Hong and Sheng Fan and Xiao Tang},
  title = {{OpenManus: An Open-Source Framework for Building General AI Agents}},
  year = {2025},
  publisher = {Zenodo},
  doi = {10.5281/zenodo.15186407},
  url = {https://doi.org/10.5281/zenodo.15186407},
}

@article{tao2025webshaper,
  title={Webshaper: Agentically data synthesizing via information-seeking formalization},
  author={Tao, Zhengwei and Wu, Jialong and Yin, Wenbiao and Zhang, Junkai and Li, Baixuan and Shen, Haiyang and Li, Kuan and Zhang, Liwen and Wang, Xinyu and Jiang, Yong and others},
  journal={arXiv preprint arXiv:2507.15061},
  year={2025}
}

@article{qiu2025alita,
  title={Alita: Generalist agent enabling scalable agentic reasoning with minimal predefinition and maximal self-evolution},
  author={Qiu, Jiahao and Qi, Xuan and Zhang, Tongcheng and Juan, Xinzhe and Guo, Jiacheng and Lu, Yifu and Wang, Yimin and Yao, Zixin and Ren, Qihan and Jiang, Xun and others},
  journal={arXiv preprint arXiv:2505.20286},
  year={2025}
}

@article{zhou2025memento,
  title={Memento: Fine-tuning llm agents without fine-tuning llms},
  author={Zhou, Huichi and Chen, Yihang and Guo, Siyuan and Yan, Xue and Lee, Kin Hei and Wang, Zihan and Lee, Ka Yiu and Zhang, Guchun and Shao, Kun and Yang, Linyi and others},
  journal={Preprint},
  year={2025}
}

@article{chai2025scimaster,
  title={SciMaster: Towards General-Purpose Scientific AI Agents, Part I. X-Master as Foundation: Can We Lead on Humanity's Last Exam?},
  author={Chai, Jingyi and Tang, Shuo and Ye, Rui and Du, Yuwen and Zhu, Xinyu and Zhou, Mengcheng and Wang, Yanfeng and Zhang, Yuzhi and Zhang, Linfeng and Chen, Siheng and others},
  journal={arXiv preprint arXiv:2507.05241},
  year={2025}
}

@article{yuan2024craft,
  title={CRAFT: Customizing LLMs by Creating and Retrieving from Specialized Toolsets},
  author={Yuan, Lifan and Phan, Hai and Chen, Yangyi and Zhang, Hongzhi and Yao, Rui and Li, Yunzhu and Ji, Heng},
  journal={arXiv preprint arXiv:2309.17428},
  year={2024}
}

@article{qian2023creator,
  title={CREATOR: Disentangling Abstract and Concrete Reasonings of Large Language Models through Tool Creation},
  author={Qian, Cheng and Han, Chi and Fung, Yi R and Qin, Yujia and Liu, Zhiyuan and Ji, Heng},
  journal={arXiv preprint arXiv:2305.14318},
  year={2023}
}

@article{cai2024toolmakers,
  title={Large Language Models as Tool Makers},
  author={Cai, Tianle and Wang, Xuezhi and Ma, Tengyu and Chen, Xinyun and Zhou, Denny},
  journal={arXiv preprint arXiv:2305.17126},
  year={2024}
}

@article{schick2023toolformer,
  title={Toolformer: Language models can teach themselves to use tools},
  author={Schick, Timo and Dwivedi-Yu, Jane and Dess{\`\i}, Roberto and Raileanu, Roberta and Lomeli, Maria and Hambro, Eric and Zettlemoyer, Luke and Cancedda, Nicola and Scialom, Thomas},
  journal={Advances in Neural Information Processing Systems},
  volume={36},
  pages={68539--68551},
  year={2023}
}

@article{wolflein2025llm,
  title={Llm agents making agent tools},
  author={W{\"o}lflein, Georg and Ferber, Dyke and Truhn, Daniel and Arandjelovi{\'c}, Ognjen and Kather, Jakob Nikolas},
  journal={arXiv preprint arXiv:2502.11705},
  year={2025}
}

@article{li2025cross,
  title={Cross-Task Experiential Learning on LLM-based Multi-Agent Collaboration},
  author={Li, Yilong and Qian, Chen and Xia, Yu and Shi, Ruijie and Dang, Yufan and Xie, Zihao and You, Ziming and Chen, Weize and Yang, Cheng and Liu, Weichuan and others},
  journal={arXiv preprint arXiv:2505.23187},
  year={2025}
}

@article{yang2024cops,
  title={Cops: Empowering llm agents with provable cross-task experience sharing},
  author={Yang, Chen and Zhao, Chenyang and Gu, Quanquan and Zhou, Dongruo},
  journal={arXiv preprint arXiv:2410.16670},
  year={2024}
}

@article{sumers2023cognitive,
  title={Cognitive architectures for language agents},
  author={Sumers, Theodore and Yao, Shunyu and Narasimhan, Karthik and Griffiths, Thomas},
  journal={Transactions on Machine Learning Research},
  year={2023}
}

@article{shinn2024reflexion,
  title={Reflexion: Language agents with verbal reinforcement learning},
  author={Shinn, Noah and Cassano, Federico and Gopinath, Ashwin and Narasimhan, Karthik and Yao, Shunyu},
  journal={Advances in Neural Information Processing Systems},
  volume={36},
  pages={8634--8652},
  year={2023}
}

@article{packer2023memgpt,
  title={MemGPT: Towards LLMs as Operating Systems},
  author={Packer, Charles and Fang, Vivian and Patil, Shishir G and Lin, Kevin and Wooders, Sarah and Gonzalez, Joseph E},
  journal={arXiv preprint arXiv:2310.08560},
  year={2023}
}

@article{chhikara2025mem0,
  title={Mem0: Building production-ready ai agents with scalable long-term memory},
  author={Chhikara, Prateek and Khant, Dev and Aryan, Saket and Singh, Taranjeet and Yadav, Deshraj},
  journal={arXiv preprint arXiv:2504.19413},
  year={2025}
}

@article{wang2023enhancing,
  title={Enhancing large language model with self-controlled memory framework},
  author={Wang, Bing and Liang, Xinnian and Yang, Jian and Huang, Hui and Wu, Shuangzhi and Wu, Peihao and Lu, Lu and Ma, Zejun and Li, Zhoujun},
  journal={arXiv preprint arXiv:2304.13343},
  year={2023}
}

@inproceedings{mialon2023gaia,
  title={Gaia: a benchmark for general ai assistants},
  author={Mialon, Gr{\'e}goire and Fourrier, Cl{\'e}mentine and Wolf, Thomas and LeCun, Yann and Scialom, Thomas},
  booktitle={The Twelfth International Conference on Learning Representations},
  year={2023}
}

@article{lewis2020retrieval,
  title={Retrieval-augmented generation for knowledge-intensive nlp tasks},
  author={Lewis, Patrick and Perez, Ethan and Piktus, Aleksandra and Petroni, Fabio and Karpukhin, Vladimir and Goyal, Naman and K{\"u}ttler, Heinrich and Lewis, Mike and Yih, Wen-tau and Rockt{\"a}schel, Tim and others},
  journal={Advances in neural information processing systems},
  volume={33},
  pages={9459--9474},
  year={2020}
}

@article{park2023generative,
  title={Generative agents: Interactive simulacra of human behavior},
  author={Park, Joon Sung and O'Brien, Joseph C and Cai, Carrie J and Morris, Meredith Ringel and Liang, Percy and Bernstein, Michael S},
  journal={arXiv preprint arXiv:2304.03442},
  year={2023}
}

@article{yao2022react,
  title={React: Synergizing reasoning and acting in language models},
  author={Yao, Shunyu and Zhao, Jeffrey and Yu, Dian and Du, Nan and Shafran, Izhak and Narasimhan, Karthik and Cao, Yuan},
  journal={arXiv preprint arXiv:2210.03629},
  year={2022}
}

@article{weston2014memory,
  title={Memory networks},
  author={Weston, Jason and Chopra, Sumit and Bordes, Antoine},
  journal={arXiv preprint arXiv:1410.3916},
  year={2014}
}

@article{chen2025sets,
  title={Sets: Leveraging self-verification and self-correction for improved test-time scaling},
  author={Chen, Jiefeng and Ren, Jie and Chen, Xinyun and Yang, Chengrun and Sun, Ruoxi and Yoon, Jinsung and Ar{\i}k, Sercan {\"O}},
  journal={arXiv preprint arXiv:2501.19306},
  year={2025}
}

@inproceedings{cormack2009reciprocal,
  title={Reciprocal rank fusion outperforms condorcet and individual rank learning methods},
  author={Cormack, Gordon V and Clarke, Charles LA and Buettcher, Stefan},
  booktitle={Proceedings of the 32nd international ACM SIGIR conference on Research and development in information retrieval},
  pages={758--759},
  year={2009}
}

@misc{damodaran2024splade,
  author={Damodaran, P.},
  title={Splade\_PP\_en\_v2: Independent Implementation of SPLADE++ Model (a.k.a splade-cocondenser* and family) for the Industry setting},
  year={2024},
  howpublished={\url{https://huggingface.co/prithivida/Splade_PP_en_v2}}
}

@article{bengio2009curriculum,
  title={Curriculum learning},
  author={Bengio, Yoshua and Louradour, J{\'e}r{\^o}me and Collobert, Ronan and Weston, Jason},
  journal={Proceedings of the 26th annual international conference on machine learning},
  pages={41--48},
  year={2009}
}
